\documentclass{article}
\usepackage[british]{babel}
\usepackage{jfrExamplee}
\usepackage{graphicx}
\usepackage{setspace}
\usepackage{apacite}
\usepackage{natbib}
\usepackage{epstopdf}
\usepackage{subcaption}
\usepackage{apalike}

\usepackage{amsmath}
\usepackage{amssymb}
\usepackage{booktabs}
\usepackage{xcolor}
\usepackage{multicol}
\usepackage{multirow}

\usepackage{booktabs}
\usepackage{array}

\newcolumntype{C}[1]{>{\centering\arraybackslash}p{#1}}

\usepackage{algorithm}
\usepackage{algorithmic}

\usepackage{endnotes}
\let\footnote=\endnote

\usepackage{lineno}

\title{A Modular Dual-Arm Apple Harvesting Robot with Enhanced Field Performance }

\author{
	Keyi Zhu \\
	Department of Mechanical Engineering \\ 
	Michigan State University \\ 
	East Lansing, MI 48824, USA \\
	\texttt{zhukeyi1@msu.edu} \\
	\And
	Kyle Lammers \\
	Department of Electrical and Computer Engineering \\ 
	Michigan State University \\ 
	East Lansing, MI 48824, USA \\
	\texttt{lammer18@msu.edu}
	\And	
        Chaaran Arunachalam \\
	Department of Electrical and Computer Engineering \\ 
	Michigan State University \\ 
	East Lansing, MI 48824, USA \\
	\texttt{arunach6@msu.edu}
	\And
	Kaixiang Zhang \\
	Department of Mechanical Engineering \\ 
	Michigan State University \\ 
	East Lansing, MI 48824, USA \\
	\texttt{zhangk64@msu.edu} \\
	\And
	Renfu Lu \\
	United States Department of Agriculture \\ 
	Agricultural Research Service \\
	East Lansing, MI 48824, USA \\
	\texttt{renfu.lu@usda.gov} \\
	\And
	Zhaojian Li\thanks{Zhaojian Li is the corresponding author.} \\
	Department of Mechanical Engineering \\ 
	Michigan State University \\ 
	East Lansing, MI 48824, USA \\
	\texttt{lizhaoj1@egr.msu.edu} \\
}

\begin{document}
\maketitle

\doublespacing

 \vspace{-16pt}
\begin{abstract}
\vspace{-14pt}
Robotic apple harvesting offers a promising solution to growing labor shortages in commercial orchards, but low throughput and poor performance in challenging orchard environments hinder its commercial adoption. This paper presents a modular dual-arm apple harvesting robot that uses a vertically stacked arm configuration to enable simultaneous operation in the upper and lower canopy zones of a single tree, simplifying platform positioning from multi-tree lateral repositioning to single-tree stops. Compared to our prior horizontal dual-arm system, the platform integrates five coordinated advances: (1)~a foundation-model-based perception pipeline combining Grounding-DINO and EfficientViT-SAM for robust fruit localization in unstructured outdoor environments; (2)~7th-order jerk-bounded trajectory generation paired with a Control Barrier Function safety filter to achieve fast yet safe arm motions; (3)~a linear sweep harvesting strategy with a 10\,cm approach buffer and rotational detachment that improves picking reliability; (4)~a temporal-logic-based dual-arm coordination policy with vision--arm asynchronous scheduling that maximizes utilization of a shared vacuum source; and (5)~field validation in two commercial orchards covering Nicoter and Envy apple varieties with different tree architectures during the 2025 harvest season. Indoor coordination experiments demonstrated a system throughput of 3.65\,s per attempt, achieving a 1.76$\times$ speedup over serialized operation while incurring only 3.7\% overhead relative to an unconstrained dual-pump upper bound. Across the 1{,}738 arm cycles collected in these field trials, the system achieved an 80.0\% per-attempt success rate and a mean per-arm cycle time of 7.53\,s. Fruit damage assessments confirmed that 91.2\% of robotically harvested fruit retained the highest USDA grade (Extra Fancy), with bruise rates between 2.4\% and 4.9\%. With further improvements in the picking cycle time and handling of heavy foliage occlusions, this new modular robot design holds promise for commercial harvesting of apples.

\end{abstract}

\section{Introduction}\label{sec:intro}

Apple production is an important segment of the U.S. agricultural economy, with harvesting being the most labor-intensive operation.
In Washington State, labor accounts for approximately 27.9\% of total apple production costs for premium fresh-market varieties such as Honeycrisp \citep{gallardo20202019}, and persistent shortages of seasonal harvest workers have compounded the economic pressure on growers.
Mechanized approaches such as targeted shake-and-catch systems have been explored as a means of reducing labor demand \citep{zhang2020field}, but these methods are unsuitable for fresh market apples, which require individual handling to prevent bruising and maintain quality grade.
This motivates the development of robotic systems capable of fruit-by-fruit harvesting in commercial orchard environments.

Robotic apple harvesting has advanced considerably over the past decade.
\citet{silwal2017design} developed one of the early comprehensive systems, integrating a custom manipulator with a vacuum-based end-effector and validating it in field trials.
\citet{bu2022design} investigated optimized picking motion patterns to improve detachment reliability.
Our group has developed different versions of apple harvesting robots from a 3-DOF single-arm platform with a 52.1\% field success rate \citep{zhang2022algorithm} to a 4-DOF system achieving 65.2--82.4\% success across orchards of varying canopy complexity \citep{zhang2024automated}.
Broader reviews of recent progress in robotic harvesting are available in \citet{rajendran2024towards} and \citet{zhou2022intelligent}.
Despite these advances, single-arm systems are fundamentally limited by their sequential operation: only one fruit can be pursued at a time, which limits overall harvesting throughput regardless of individual cycle time improvements.

Reliable fruit localization in unstructured orchard environments remains a core challenge for all of the above systems.
Deep learning-based methods—including convolutional neural networks for apple detection \citep{chu2021deep} and instance segmentation for pixel-level fruit identification \citep{williams2019robotic}—have substantially improved detection accuracy for a range of crops \citep{suresh2023selective, yang2023vision}.
However, accurate 3D localization has proven more difficult than detection: as surveyed by \citet{fu2020application}, consumer-grade RGB-D cameras yield unreliable depth measurements under harsh outdoor lighting and partial occlusion, making the gap between detection and localization a persistent bottleneck.
More recently, vision foundation models—including the Segment Anything Model (SAM) \citep{kirillov2023segment}, Grounding-DINO \citep{liu2024groundingdino}, and EfficientViT-SAM \citep{zhang2024efficientvit}—have demonstrated strong zero-shot and few-shot generalization, opening new possibilities for robust perception in agricultural settings.
In apple harvesting, these models have been applied to both 3D localization through foundation-model-based detection combined with depth clustering \citep{bhattacharya2025detect}, and to ripeness and size estimation for selective harvesting \citep{zhu2025foundation}.

Multi-arm configurations have been explored as a direct path to higher throughput.
\citet{yoshida2022automated} and \citet{gursoy2023towards} demonstrated dual-arm apple-harvesting platforms, though without structured inter-arm coordination.
For other crops, \citet{barnett2020work} studied work distribution among Cartesian kiwifruit-harvesting arms, and \citet{arikapudi2023robotic} analyzed cycle time trade-offs in multi-arm array systems.
In apple harvesting, \citet{li2023multi} developed a four-arm platform with reinforcement-learning-based task coordination \citep{li2023RL}, and our group introduced dual-arm apple harvesting~\citep{lammers2024development}
and subsequently advanced it to achieve an 80.7\% success rate, reducing harvest time by 28\% compared to a single-arm baseline~\citep{zhu2025advancement}.

More recent work has begun to address the coordination strategies that govern multi-arm operation.
\citet{yan2025intelligent} proposed a workspace zoning method that partitions the canopy between two arms by spatial position, achieving up to 72.4\% parallel operation in orchard experiments.
\citet{guo2025dynamic} applied an LSTM-PPO reinforcement learning framework to online task re-planning for multi-arm systems, demonstrating 8--11\% cycle time reductions over a genetic algorithm baseline in field tests.
Despite this progress, key challenges remain: existing coordination strategies either rely on simplified spatial partitioning that does not exploit inter-arm timing opportunities, or require substantial online computation.
More broadly, safety guarantees for high-speed dual-arm motion and tight coupling between perception latency and arm scheduling have not been systematically addressed.

This paper presents an improved dual-arm apple harvesting robot that addresses these limitations through coordinated advances in hardware design, perception, motion control, and system-level coordination.
Compared to our prior system \citep{zhu2025advancement}, the new vertically stacked dual-arm system  introduces the following key advancements:

\begin{itemize}
    \item \textbf{Foundation-model-based perception:} A fine-tuned EfficientViT-SAM replaces the custom segmentation model used in prior work, providing improved segmentation quality and generalization while maintaining real-time deployment capability.
    \item \textbf{Jerk-bounded trajectory with formal safety:} Seventh-order polynomial trajectories with explicit jerk limits are paired with a Control Barrier Function (CBF) safety filter, providing formal guarantees that joint motion remains within safe operating limits during high-speed operation.
    \item \textbf{Linear sweep harvesting strategy:} A two-stage motion design with a 10~cm approach buffer and a 1.0~s linear sweep phase improves detachment reliability for apples with limited direct-approach access.
    \item \textbf{Temporal-logic-based dual-arm coordination with vision-arm asynchrony:} A formal temporal-logic specification governs the shared-vacuum mutex and failure-recovery branching between the two arms, while an asynchronous vision--arm scheduling layer triggers early arm handoff at an intermediate via-point, saving approximately 1~s per harvest cycle compared to sequential handoff.
    \item \textbf{Field validation in commercial orchards:} The complete system was evaluated in two commercial orchards during the 2025 harvest season in Washington State, for Nicoter and Envy apple varieties under different tree architectures.
\end{itemize}

The remainder of this paper is organized as follows.
Section~\ref{sec:hardware} describes the hardware system design.
Section~\ref{sec:software} presents the software architecture, including perception, motion control, and coordination.
Section~\ref{sec:experiment} reports field trial results and analysis.
Section~\ref{sec:discussion} discusses system limitations and future directions, and Section~\ref{sec:conclusion} concludes the paper.

\section{Hardware System}\label{sec:hardware}
\begin{figure}[!t]
    \centering
    \includegraphics[width=0.65\linewidth]{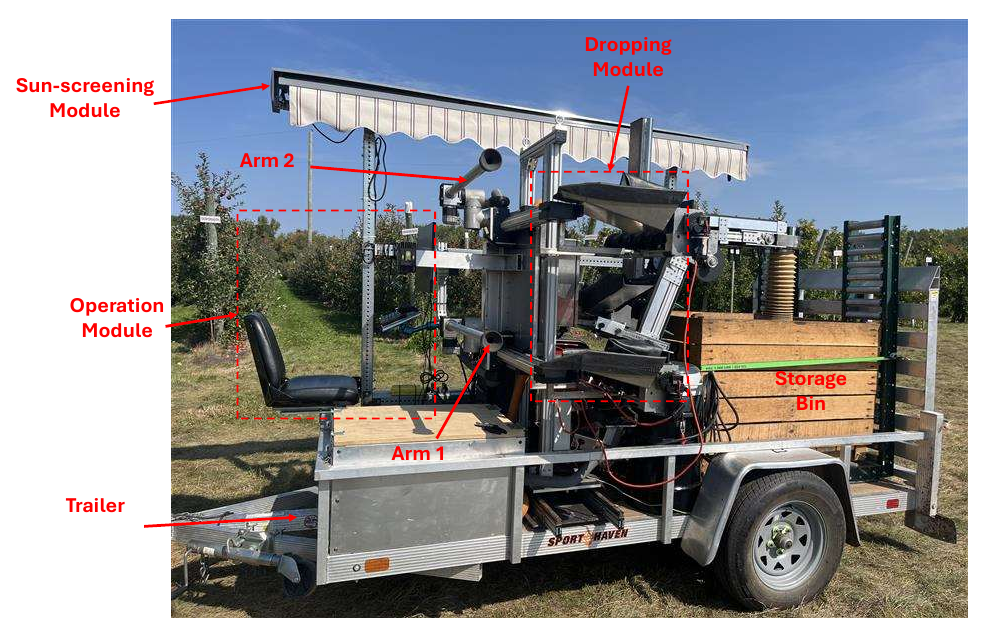}
    \caption{Overview of the dual-arm apple harvesting robot mounted on a trailer platform.}
    \label{fig:robot_overview}
\end{figure}

The robotic harvesting platform is mounted on a trailer for deployment in commercial orchards and transported by an agricultural tractor, following a stop-and-go harvesting pattern under human operator control. As shown in Figure~\ref{fig:robot_overview}, the platform integrates four hardware modules: a perception component, a vertically-configured dual-arm harvester, a shared vacuum and valve unit, and a post-harvest fruit-handling system. In addition to these core modules, several auxiliary subsystems support sustained field operation. Power is supplied by a Honda gas-powered generator that provides 240~V at 5.5~kW and can run the entire system for over five hours on a full tank. Onboard computation is split between two computers: a perception workstation---an industrial computer with a 24-core Intel Core\textsuperscript{\textregistered} i9-13900E CPU, 64~GiB of RAM, and an NVIDIA RTX A6000 GPU (48~GiB)---that runs the foundation-model-based perception pipeline, and a real-time control computer (NVIDIA Jetson AGX Orin 64GB) that executes trajectory generation, safety filtering, motion control, and dual-arm coordination. An operator module, consisting of a monitor, keyboard, and mouse, is mounted on the trailer for system setup, monitoring, and manual intervention during field operation. To improve perception robustness in outdoor environments, a retractable sun-screening module is installed above the workspace to block direct sunlight, thereby mitigating specular reflections and overexposure. 

Two key hardware modifications distinguish this system from our previous dual-arm platform~\citep{zhu2025advancement}. First, the arm layout has been redesigned from a horizontal side-by-side configuration to a vertical stacking arrangement. In the previous system, the two arms operated across overlapping regions of adjacent trees, leading to sensitivity to inter-tree spacing, incomplete coverage of individual trees, and frequent platform height adjustments to fully harvest one tree. The present vertical layout mounts the two arms one above the other, directing both at the same tree simultaneously. This concentrates harvesting effort on a single tree per platform stop, reducing repositioning frequency and improving overall throughput. Second, the pan and tilt joints of each manipulator have been upgraded to ZeroErr servo motors with CAN bus communication, replacing the motors used in the previous system and improving actuation reliability under field conditions.

\subsection{RGB-D Sensing Module}\label{sec:perception_hardware}
Reliable fruit detection and localization relies on a stable RGB-D sensing setup. Because both arms operate within a shared workspace simultaneously, the sensing module should maintain uninterrupted coverage without being affected by arm motion. To this end, the system employs a TL460-S1-E1 Time-of-Flight (ToF) camera (Percipio Inc., Shanghai, China; $62^\circ \times 49^\circ$ FOV, 0.3--9.5~m range). The camera is mounted on the robot frame approximately 1~m from the front boundary of the manipulator workspace. The camera delivers synchronized $1920 \times 1080$ RGB and $640 \times 480$ depth frames with a depth accuracy of $\pm 4$~mm plus 0.25\% of the measured range. This frame-fixed arrangement provides a stable viewpoint that spans the full reachable workspace of both manipulators, while avoiding the motion blur and depth degradation that arise when it comes to eye-in-hand configuration.

Two primary challenges affect perception in outdoor orchard environments. The first is extreme lighting conditions, including direct sunlight and specular reflections that cause overexposure and degrade image quality. This is mitigated by the retractable sun-screening module installed on top of the trailer, which can be extended before harvesting begins to block direct sunlight from reaching the canopy. The second challenge is arm-induced occlusion: the manipulator arms regularly traverse the camera's field of view during operation, temporarily occluding portions of the canopy and degrading depth measurements in the affected region. This is a persistent challenge in eye-to-hand sensing configurations. In this work, we mitigate it through a view-management strategy detailed in Section~\ref{sec:vision_arm_coord}.

\subsection{Vertically-Configured Dual-Arm Harvester}\label{sec:dual_arm_hardware}

Each manipulator is a 4-degree-of-freedom (DOF) serial chain consisting of one prismatic joint (linear stage) and three revolute joints for pan, tilt, and end-effector rotation. The arm body is a hollow tube that serves as the vacuum airflow channel: when connected to the centralized vacuum source (Section~\ref{sec:vacuum_system}), suction is routed through the tube interior directly to the end-effector, eliminating the need for external hoses and maintaining a compact structure. The fourth DOF enables $360^\circ$ rotation of the end-effector about the tube axis, allowing the arm to twist and break the stem during detachment without repositioning the entire manipulator. This twisting action reduces interference with surrounding branches and improves detachment reliability in dense canopy regions.

The end-effector at the tip of each arm needs to form an airtight seal against the curved apple surface without causing bruising. To meet this requirement, each end-effector is cast from 40~Shore~A silicone rubber, soft enough to conform to the fruit surface while maintaining sufficient stiffness to sustain the vacuum pressure differential. A 3D-printed connector secures the silicone cup to the arm tube, providing a rigid, slip-free interface between the compliant contact surface and the structural airflow channel. When connected to the Delfin vacuum source, the resulting suction force averages 45~N, which is sufficient for reliable attachment and detachment under field conditions. Laboratory tests show that the suction cup can establish attachment when its center is offset from the apple center by up to 1.5~cm, a margin that comfortably absorbs the localization uncertainty typical of outdoor ToF measurements. This tolerance also simplifies the perception requirement to a single 3D centroid estimate per fruit, eliminating the need for detailed orientation or shape estimation required by mechanical grippers.
The vacuum approach further reduces fruit bruising and is inherently robust to variations in apple orientation.

\begin{figure}[!htbp]
    \centering
    \includegraphics[width=0.4\linewidth]{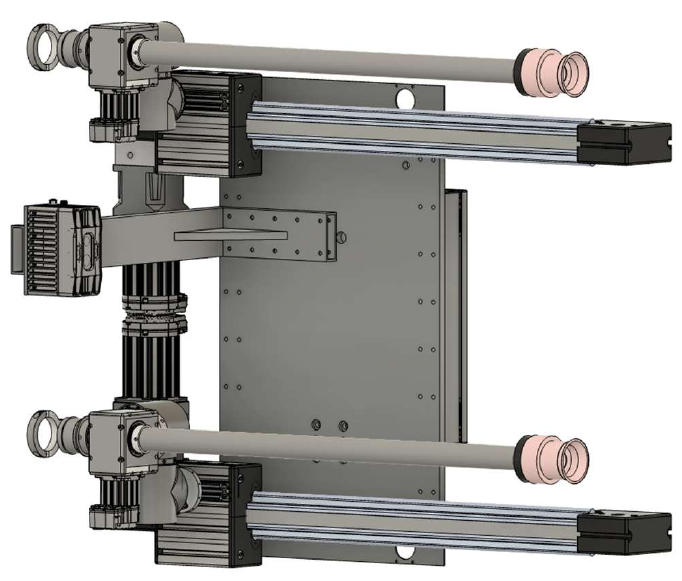}
    \caption{CAD rendering of the vertically-configured dual-arm harvester.}
    \label{fig:hardwares}
\end{figure}

The two manipulators are arranged in a vertical stacking configuration, as shown in Figure~\ref{fig:hardwares} and labeled Arm~1 and Arm~2 in Figure~\ref{fig:robot_overview}. The upper and lower arms are mounted on separate support frames with a center-to-center spacing of 60~cm. This configuration partitions the canopy vertically: the upper arm (Arm~2) primarily serves the upper fruiting region, while the lower arm (Arm~1) targets the lower region. A small overlap is maintained between the two workspaces so that fruit near the boundary can still be assigned without leaving gaps in coverage. 
Apples in this overlap region are allocated based on vertical position, with higher fruit assigned to the upper arm and lower fruit to the lower arm.
This strategy preserves full canopy coverage while reducing workload imbalance between the two arms.

The linear stage and end-effector rotation joint of each manipulator are driven by Teknic DC servo motors with integrated controllers and encoders, communicating with the onboard control computer (NVIDIA Jetson AGX Orin 64GB) through a Teknic USB hub. The pan and tilt joints use ZeroErr servo motors communicating over CAN bus at 100~Hz, replacing the USB-based motors used in our previous system~\citep{zhu2025advancement}. CAN bus provides deterministic timing and higher noise immunity compared to serial communication, improving actuation reliability under the electrically noisy field conditions produced by the generator, vacuum motor, and valve actuators operating simultaneously.

To maintain reliable operation under this mixed communication architecture, a watchdog process runs in ROS2 and monitors encoder updates from both the CAN and USB links at 100~Hz. If a joint fails to report within 500~ms or if encoder readings exhibit an anomalous jump, the system triggers an emergency stop. Recovery is performed by manually moving the manipulators to a safe configuration and then executing an automated re-homing sequence. Although future versions will migrate toward a unified communication architecture, the present implementation has demonstrated sufficient robustness during extended orchard trials.

To accommodate varying orchard structures, a platform movement module enables repositioning of the harvester on the trailer. The module is actuated by two stepper motors and provides motion in both horizontal and vertical directions, communicating with the onboard computer through an Arduino Uno board. In addition to the software-based control available through the graphical user interface, the current system introduces a dedicated physical button panel mounted near the operator position. This allows the operator to reposition the platform quickly during field operation without interacting with the computer, reducing setup time at each harvesting stop and improving usability under practical working conditions.

\subsection{Centralized Vacuum and Airflow Distribution}\label{sec:vacuum_system}
Both manipulators share a single centralized vacuum source rather than each arm having its own pump. The system uses a Delfin industrial vacuum machine (model 202 DS), rated at 5.5~HP, with suction pressure up to 2500~mmH$_2$O and airflow rate of 360~m$^3$/h. The machine operates at full power with continuous airflow throughout harvesting. 
A dedicated vacuum pump per arm would sit idle whenever that arm is not actively attaching or detaching fruit, wasting energy during a substantial portion of each picking cycle. However, optimizing the vacuum power sharing between the two arms presents significant control complexity. With a shared source, one or both arms draw suction at any given time, so the machine's output is utilized nearly continuously and the system runs at a single fixed power level without dynamic power management. The Delfin vacuum source connects directly to a custom valve unit that selectively routes suction to either manipulator or vents a manipulator to the atmosphere for fruit release.

\begin{figure}[!htbp]
    \centering
    \begin{subfigure}[b]{0.3\textwidth}
        \centering
        \includegraphics[width=\linewidth]{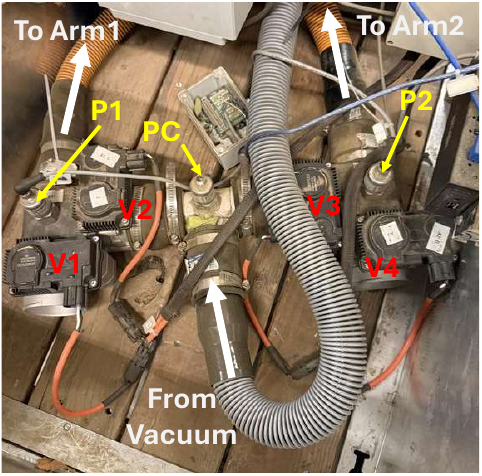}
        \caption{}
        \label{fig:valve_photo}
    \end{subfigure}
    \hspace{0.05\textwidth}
    \begin{subfigure}[b]{0.85\textwidth}
        \centering
        \includegraphics[width=\linewidth]{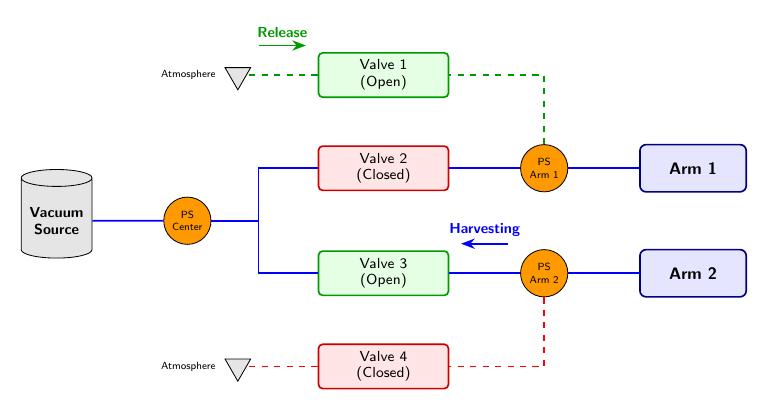}
        \caption{}
        \label{fig:valve_schematic_sub}
    \end{subfigure}
    \caption{The shared vacuum and valve unit. (a)~Physical assembly showing the four butterfly-style gate valves (V1--V4), pressure sensors (PS) for Arm~1 (P1), Arm~2 (P2), and the central inlet (PC), with hose connections to the vacuum source and both arms. (b)~Schematic of the airflow routing: vacuum can be directed to either arm for attachment, while the other arm is vented to the atmosphere for fruit release.}
    \label{fig:valve_schematic}
\end{figure}

The valve unit, shown in Figure~\ref{fig:valve_schematic}, contains four butterfly-style gates driven by compact 12~V DC motors. Valves~1 and~2 control the airflow to Arm~1: one connects the arm channel to the vacuum source for attachment, while the other vents it to the atmosphere for fruit release. Valves~3 and~4 serve the same functions for Arm~2. By adjusting the gate states independently, each arm can transition between harvesting and releasing modes without affecting the other. Each valve transitions in less than one second, enabling rapid switching during coordinated dual-arm operation. Because the valve motors run continuously throughout each harvesting session, a pulse-width-modulation drive scheme limits their average power draw and prevents thermal buildup in the compact valve enclosure. The valve unit communicates with the onboard computer through an Arduino Uno microcontroller using simple digital commands. To monitor system state, pressure sensors are installed at three key locations---one in each arm channel (P1 and P2) and one at the vacuum inlet (PC)---and sampled at 100~Hz. By comparing readings across the three sensors, the system can determine whether an apple has been successfully attached to the end-effector, whether suction is maintained during retraction, or whether attachment has been lost. The detailed logic for translating these pressure signals into coordination decisions is described in Section~\ref{sec:coordination}.

\subsection{Post-Harvest Fruit Handling}\label{sec:fruit_handling}

\begin{figure}[htbp]
    \centering
    \includegraphics[width=0.5\linewidth]{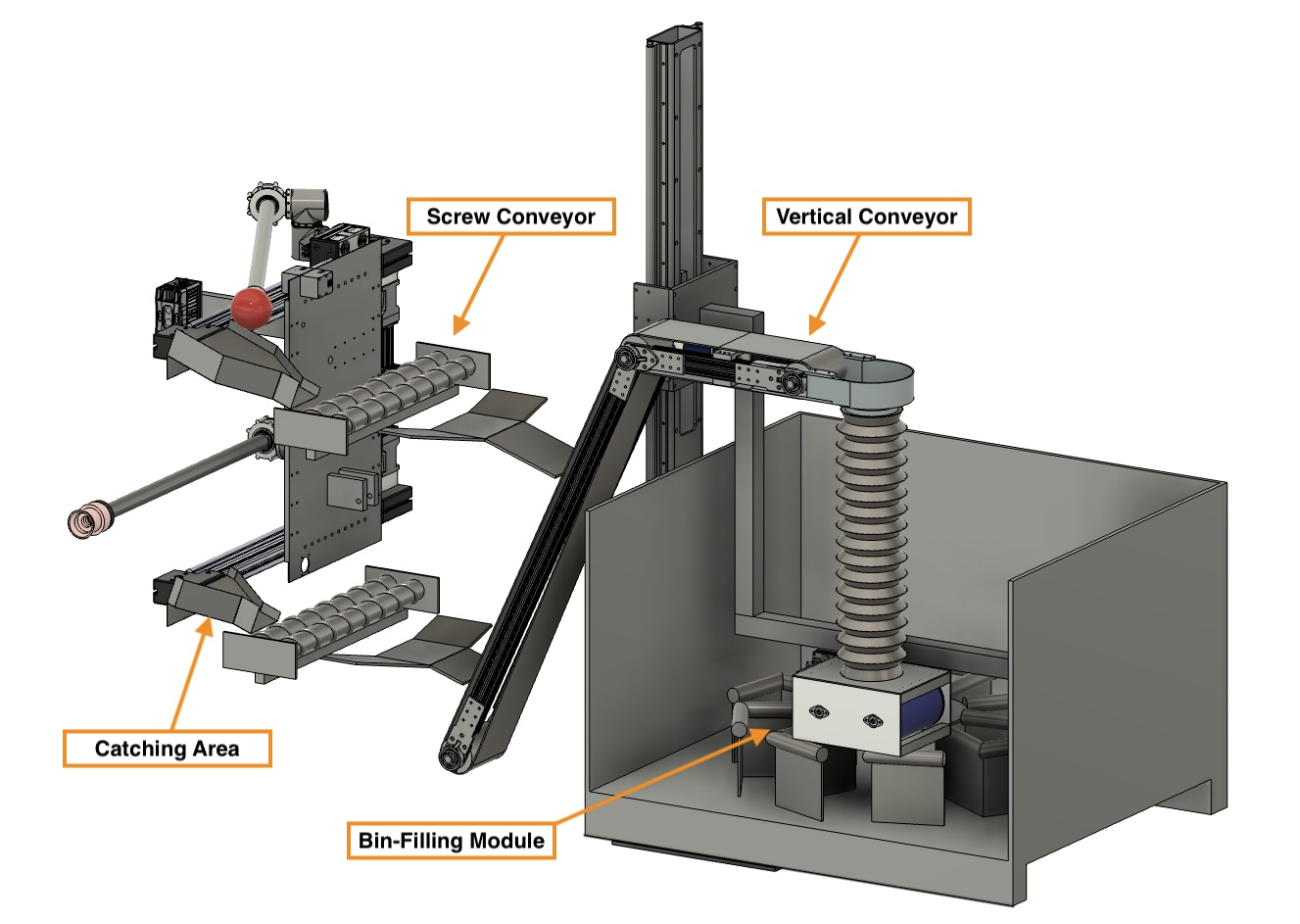}
    \caption{CAD rendering of the post-harvest fruit-transport path. After an apple is released by a manipulator, it is received at the \emph{catching area} (the per-arm dropping modules and food-grade cushioned ramps), carried laterally by the \emph{screw conveyor}, lowered into the storage bin by the \emph{vertical conveyor}, and finally spread across the bin by the \emph{bin-filling module}. All fruit-contacting surfaces along the path are lined with food-grade cushioning to minimize bruising.}
    \label{fig:fruit_transport}
\end{figure}

The fruit-handling module transfers harvested apples from the manipulators to a storage bin while minimizing bruising. All surfaces that contact the fruit throughout the harvest process---from the silicone end-effector to the dropping modules, ramps, conveyors, and bin-filling components---are constructed from or lined with food-grade materials. Unlike the previous system~\citep{zhu2025advancement}, in which the two arms shared a single dropping location, the present platform assigns one dropping module to each arm. These two modules, which together form the \emph{catching area} in Figure~\ref{fig:fruit_transport}, are mounted on the left side of the linear stages and inclined at approximately $20^\circ$. Each dropping surface is lined with soft foam that cushions the impact when an apple is released from the end-effector, reducing bruising before the fruit enters the downstream transport path. This lateral placement decouples fruit deposition from the main picking workspace and reduces interference between the manipulators and the collection hardware during downward harvesting motions.

The offset geometry of the dropping modules, however, introduces a repeatable motion constraint: the arms must rotate outward before reaching the release position. To keep these motions predictable, all release trajectories are routed through a fixed via-point located 5~cm to the right of the release position. In this way, the hardware geometry is translated into an explicit trajectory constraint, while the implementation details of the motion policy are handled in Section~\ref{sec:motion_control}.

After release, each apple travels through a staged transport path---the screw conveyor and vertical conveyor in Figure~\ref{fig:fruit_transport}---that carries it from the catching area to the storage bin while keeping impacts low: all contact surfaces are lined with food-grade cushioning, and the path is arranged to limit drop heights at each transfer. At the end of this path, the bin-filling module decelerates each apple and distributes the fruit evenly across the bin to prevent collisions between incoming fruit and those already collected, following the design principles established in our prior work~\citep{zhang2019improvements, lu2022development}. The vertical position of the bin filler is adjusted as the bin fills to maintain a short drop distance, further reducing bruise risk. The entire fruit-handling system is controlled through a lightweight Arduino-based on/off interface.

\section{Software Design}\label{sec:software}
The software stack is organized into three layers. The perception pipeline (Section~\ref{sec:perception}) detects and localizes apples in 3D from the RGB-D stream. The dual-arm coordination layer (Section~\ref{sec:coordination}) builds on these detections to assign targets between the two arms, sequence each arm through its harvest cycle, and arbitrate the shared vacuum source. Each individual harvest action is then executed by the single-arm motion stack (Section~\ref{sec:motion_control}), which generates smooth approach trajectories and enforces joint-level safety.

\subsection{Foundation-Model-Based Perception}\label{sec:perception}
\begin{figure}[htbp]
    \centering
    \includegraphics[width=\linewidth]{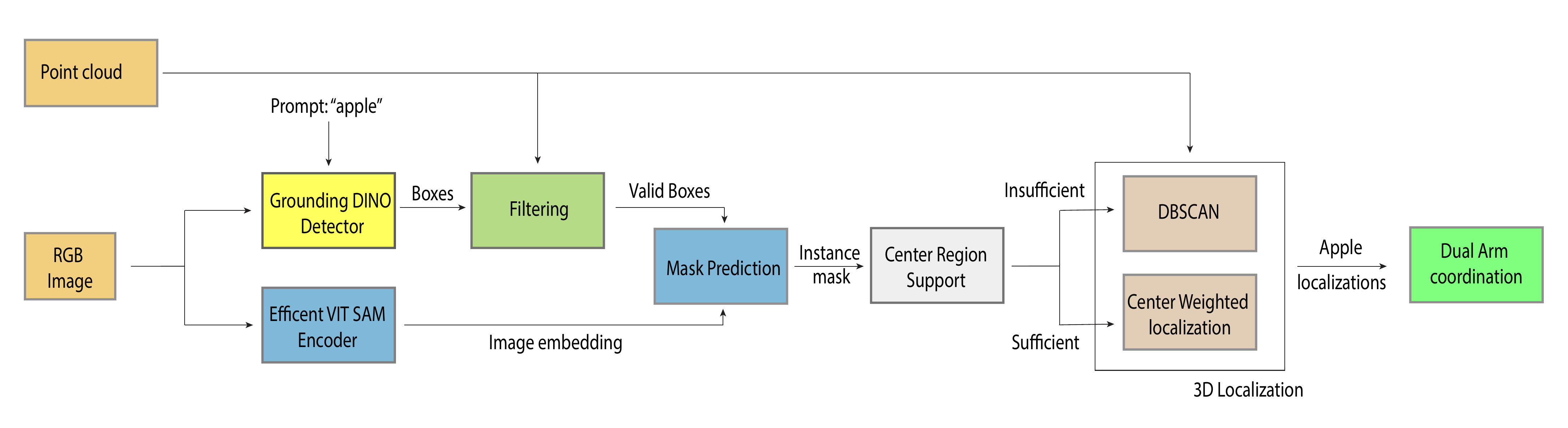}
    \caption{\textbf{Overview of the perception pipeline.} The RGB image and the registered point cloud are the inputs. An open-vocabulary detector (Grounding-DINO) prompted with the word ``apple'' produces candidate bounding boxes, which are filtered and then segmented by EfficientViT-SAM (an efficient variant of the Segment Anything Model, SAM) to obtain per-apple instance masks. Each mask is localized in three dimensions (3D): when sufficient support is available in its central region, a center-weighted depth estimate is used; otherwise, Density-Based Spatial Clustering of Applications with Noise (DBSCAN) is applied. The resulting 3D apple localizations are passed to the dual-arm coordination layer.}
    \label{fig:perception_pipeline}
\end{figure}

Accurate fruit detection and localization are essential prerequisites for arms' motion planning. To achieve robust localization under outdoor occlusion, variable lighting, and clustered fruit arrangements, the perception pipeline combines foundation-model-based 2D reasoning with depth-based geometric filtering. The pipeline consists of detection, filtering, segmentation, and 3D localization modules that progressively refine object hypotheses into actionable picking targets. This modular design balances accuracy with computational efficiency. An overview of the pipeline is shown in Figure~\ref{fig:perception_pipeline}.

\subsubsection{Dataset}\label{sec:dataset}

To evaluate the perception pipeline under deployment-representative conditions, we use a curated orchard dataset collected in Michigan. The dataset consists of 232 RGB images annotated with apple bounding boxes and instance segmentation masks, capturing variations in lighting, occlusion, and viewpoints encountered during real-world harvesting.
The dataset is grouped by acquisition date and split into disjoint training, validation, and test sets (139/45/48), with no overlap across splits. The validation and test sets are used exclusively for model selection and final evaluation.
Bounding box annotations are used for detection, while instance masks are used for segmentation. For segmentation evaluation, ground-truth bounding boxes serve as instance prompts, enabling consistent instance-conditioned evaluation across the perception pipeline.

\subsubsection{Detection}\label{sec:detection}

Modern deep learning-based object detectors have demonstrated strong performance in unstructured environments by learning robust visual representations from large-scale datasets~\citep{ren2016fasterrcnn}. In our system, we employ Grounding-DINO~\citep{liu2024groundingdino}, a Transformer-based detector that combines DETR-style detection~\citep{zhang2023dino} with grounded vision-language pretraining~\citep{li2022glip}. Consistent with our previous system, we retain this detection framework and adapt it to the current deployment setting using our internal dataset.

To improve performance under deployment-specific conditions, we use a two-stage fine-tuning strategy. Starting from a pretrained model, we first perform general adaptation on the MetaFruit dataset~\citep{metafruit2024}, followed by task-specific fine-tuning on our internal dataset. This second stage captures deployment characteristics such as partial visibility, background clutter, and lighting variation, enabling robust detection under challenging orchard conditions. The results in Table~\ref{tab:detection_training} show that both stages contribute to improved detection performance, with the second stage providing additional gains under deployment-relevant conditions.

\begin{table}[t]
\centering
\caption{\textbf{Detection performance across training stages.} 
Evaluation is performed on the internal dataset. Detection accuracy is measured using mean Average Precision at IoU $=0.5$ (mAP@0.5), capturing detection performance at a standard overlap threshold, and mAP@[.5:.95], which averages performance over multiple IoU thresholds to reflect localization quality.}
\label{tab:detection_training}
\begin{tabular}{lcc}
\toprule
\textbf{Model Stage} & \textbf{mAP@0.5} & \textbf{mAP@[.5:.95]} \\
\midrule
Pretrained Grounding-DINO & 0.750 & 0.458 \\
+ MetaFruit Fine-tuning & 0.868 & 0.586 \\
+ Internal Dataset Fine-tuning & \textbf{0.950} & \textbf{0.666} \\
\bottomrule
\end{tabular}
\end{table}

Robotic harvesting exhibits an asymmetric cost structure: missed detections prevent downstream planning and execution, whereas false positives are typically filtered by subsequent segmentation and localization stages and, at worst, result in redundant manipulation attempts. This motivates operating the detector in a recall-oriented regime, where coverage of valid targets is prioritized over early-stage precision. To realize this, we select a fixed operating threshold based on validation performance on the internal dataset: the detector confidence threshold is swept over the range $[0.0, 1.0]$ with a step size of $0.02$ to characterize precision--recall behavior, and an operating point is chosen to achieve a target recall of approximately $0.85$ at IoU = $0.5$. The final detector is operated at a confidence threshold of $0.46$, achieving a precision of $0.961$ at the selected operating point, and is used during deployment.

Operating in this recall-oriented regime produces a set of candidate detections that may include redundant or low-quality bounding boxes. To improve robustness before passing detections to downstream modules, we incorporate a lightweight filtering stage based on geometric constraints. First, redundant detections are removed by filtering boxes that are largely contained within other boxes, eliminating duplicate hypotheses for the same apple. Second, geometric constraints are applied to remove implausible detections: bounding boxes are filtered based on their size using a robust statistic computed from the median and median absolute deviation (MAD), allowing the system to reject unusually small detections while remaining stable to outliers, and an additional aspect-ratio filter rejects boxes whose shape is inconsistent with the expected image-space appearance of apples.

\begin{table}[t]
\centering
\caption{\textbf{Effect of geometric filtering at the selected operating point.} 
Precision measures the fraction of predicted detections that are correct, while Recall@0.5 measures the fraction of ground-truth objects detected at an IoU threshold of 0.5. Geometric filtering improves precision by removing structured false positives while maintaining high recall.}
\label{tab:filtering_effect}
\setlength{\tabcolsep}{6pt}
\renewcommand{\arraystretch}{1.15}
\begin{tabular}{lcc}
\toprule
\textbf{Stage} & \textbf{Precision} & \textbf{Recall@0.5} \\
\midrule
Operating Point ($\tau_{\mathrm{conf}} = 0.46$) & 0.961 & \textbf{0.84} \\
+ Geometric filtering                          & \textbf{0.972} & 0.81 \\
\bottomrule
\end{tabular}
\end{table}

The effect of geometric filtering is quantified in Table~\ref{tab:filtering_effect}. Filtering improves precision from 0.961 to 0.972 by removing structured false positives, while maintaining high recall with only a modest reduction from 0.84 to 0.81 at the selected operating point, resulting in higher-quality inputs for downstream processing. In the deployed pipeline, we additionally apply a depth-consistency check using the associated point cloud. Detections whose projected center corresponds to invalid 3D measurements (e.g., $(0,0,0)$) are removed, as they cannot be reliably localized.

\subsubsection{Instance Segmentation}\label{sec:instance_segmentation}

Following detection, each candidate apple is refined into a pixel-accurate instance mask through box-prompted segmentation, improving spatial precision for downstream 3D point extraction and localization by isolating apple pixels from surrounding foliage, branches, and background. We use the Segment Anything Model (SAM)~\citep{kirillov2023segment} as a high-capacity reference due to its strong segmentation performance, and adapt it to the target domain by fine-tuning its mask decoder on the internal dataset while keeping the image and prompt encoders fixed. Although this configuration provides high segmentation quality, the computational cost of the SAM image encoder limits its suitability for real-time deployment. To address this limitation, we adopt EfficientViT-SAM~\citep{zhang2024efficientvit}, a family of lightweight models designed for efficient segmentation while retaining the same prompting interface; EfficientViT-SAM models are initialized using feature-level distillation on the MetaFruit dataset~\cite{metafruit2024}, followed by joint optimization of the image encoder and mask decoder on the internal dataset, enabling the model to retain strong generic representations while adapting to deployment-specific characteristics.

\begin{table}[t]
\centering
\caption{\textbf{Segmentation performance and efficiency.} 
Comparison of final models showing the trade-off between segmentation quality and encoder cost. Segmentation quality is evaluated using mean Intersection-over-Union (mIoU), which measures region overlap accuracy, and boundary F1 score (BF1), which evaluates boundary alignment. Latency corresponds to the image encoder only and is reported per image using median (p50) and tail (p95) statistics. The mask predictor architecture is shared across models and is therefore not included in the comparison.}
\label{tab:segmentation_eval}
\setlength{\tabcolsep}{5pt}
\renewcommand{\arraystretch}{1.15}
\begin{tabular}{lccccc}
\toprule
\textbf{Model} & \textbf{Input Res} & \textbf{mIoU} & \textbf{BF1} & \textbf{p50 (ms)} & \textbf{p95 (ms)} \\
\midrule
SAM (Decoder FT)        & 1024 & 0.779 & 0.873 & 466.80 & 471.71 \\
EfficientViT-SAM-L0      & 512  & 0.768 & 0.860 & \textbf{9.99}   & \textbf{10.21} \\
EfficientViT-SAM-L2      & 512  & 0.783 & 0.879 & 13.67  & 14.99 \\
EfficientViT-SAM-XL1     & 1024 & \textbf{0.802} & \textbf{0.902} & 36.71  & 36.99 \\
\bottomrule
\end{tabular}
\end{table}

Table~\ref{tab:segmentation_eval} shows that EfficientViT-SAM-L2 achieves segmentation quality comparable to the SAM reference while operating at significantly lower latency. Since the mask predictor is shared across models, latency differences arise primarily from the image encoder and are reported accordingly. While the larger XL1 variant provides higher accuracy, the improvement is modest relative to its increased computational cost. EfficientViT-SAM-L2 is therefore selected for deployment as the most efficient model achieving near-reference segmentation performance.

At inference time, the image encoder processes each frame once, while the mask predictor is applied to each detected instance. To improve throughput, mask prediction is executed in batches (batch size 8), ensuring efficient GPU utilization while maintaining stable latency. Subsequently, candidate instances with fewer than a minimum number of foreground pixels are discarded, as they do not provide sufficient support for reliable 3D localization.

\begin{figure}[t]
\centering

\begin{subfigure}[t]{0.28\linewidth}
    \centering
    \begin{minipage}[c][4cm][c]{\linewidth}
        \centering
        \includegraphics[width=\linewidth]{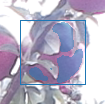}
    \end{minipage}
    \vspace{50pt}
    \caption{Occlusion}
\end{subfigure}
\hfill
\begin{subfigure}[t]{0.28\linewidth}
    \centering
    \begin{minipage}[c][4cm][c]{\linewidth}
        \centering
        \includegraphics[width=\linewidth]{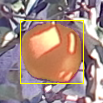}
    \end{minipage}
    \vspace{50pt}
    \caption{Lighting}
\end{subfigure}
\hfill
\begin{subfigure}[t]{0.40\linewidth}
    \centering
    \begin{minipage}[c][4cm][c]{\linewidth}
        \centering
        \includegraphics[width=\linewidth]{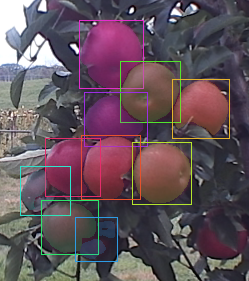}
    \end{minipage}
    \vspace{50pt}
    \caption{Dense clustering}
\end{subfigure}

\caption{\textbf{Qualitative performance under challenging conditions.}
The final detection and segmentation models demonstrate robustness to common sources of variability in orchard environments. 
(\textbf{Left}) Under occlusion from foliage and branches, the model accurately detects and segments partially visible apples. 
(\textbf{Center}) Under extreme lighting conditions, including specular highlights and overexposure, segmentation remains stable and aligned with visible regions. 
(\textbf{Right}) In densely clustered scenes with multiple overlapping fruits, the model maintains consistent instance-level separation. 
Each apple instance is shown with a distinct color for clarity, with detection boxes overlaid.}
\label{fig:qualitative_results}
\end{figure}

Figure~\ref{fig:qualitative_results} illustrates the qualitative performance of the final detection and segmentation models under challenging conditions commonly encountered in orchard environments, including occlusion, extreme lighting, and dense clustering. The models maintain consistent instance-level segmentation and detection behavior across these scenarios, supporting their suitability for deployment.

\subsubsection{3D Localization}

\begin{figure}[t]
\centering

\begin{subfigure}{0.32\linewidth}
    \centering
    \includegraphics[width=\linewidth]{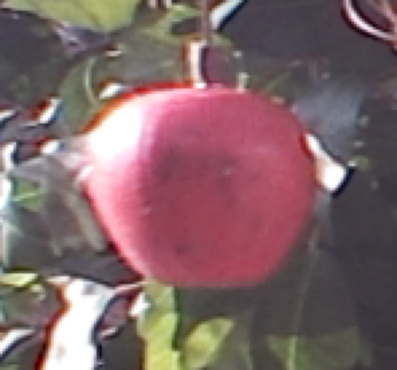}
    \caption{Detected apple}
\end{subfigure}
\hfill
\begin{subfigure}{0.32\linewidth}
    \centering
    \includegraphics[width=\linewidth]{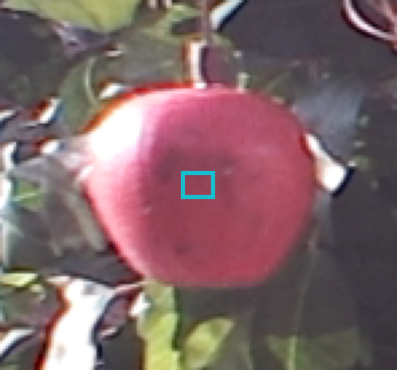}
    \caption{Central support region}
\end{subfigure}
\hfill
\begin{subfigure}{0.32\linewidth}
    \centering
    \includegraphics[width=\linewidth]{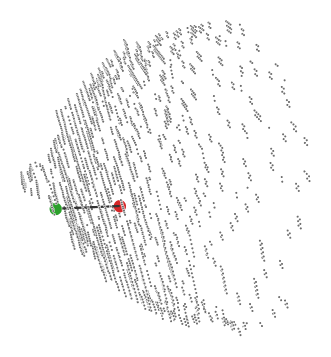}
    \caption{Localization comparison}
\end{subfigure}

\caption{\textbf{Effect of central support on localization.} 
(\textbf{Left}) Detected apple in the scene. 
(\textbf{Center}) The central region (cyan) used to assess support for center-weighted localization. 
(\textbf{Right}) Comparison of localization estimates in 3D. The center-weighted estimate (green) lies closer to the visible surface, while the DBSCAN centroid (red) is biased inward due to non-uniform point distribution. The estimated depths are 803.5\,mm (center-weighted) and 816\,mm (DBSCAN), illustrating the inward bias introduced by clustering.}
\label{fig:localization_comparison}
\end{figure}

Following segmentation, instance masks are used to extract apple-specific points from aligned RGB-D observations for 3D localization. In our previous system, localization was performed using DBSCAN~\citep{ester1996density} to cluster the segmented point cloud and estimate the apple position using the centroid of the dominant cluster. While this approach is robust to noise and occlusion, it is computationally expensive and can introduce bias in well-observed cases, where non-uniform surface sampling shifts the centroid inward from the true surface (Figure~\ref{fig:localization_comparison}). To address these limitations, we adopt a conditional localization strategy that selects the estimation method based on the reliability of the central region of the instance, motivated by the observation that clustering is unnecessary when reliable central measurements are available and that simpler estimators can provide both higher accuracy and lower computational cost in such cases.

Reliability is assessed locally at the center of the detection. A region centered within the bounding box, with size proportional to the box dimensions, is used to measure the proportion of apple pixels. If this proportion exceeds a predefined threshold, the instance is considered to have sufficient central support. For such instances, we apply center-weighted localization, estimating depth using only the points within this region. This provides a stable surface-level estimate while avoiding clustering bias and reducing computational cost. If sufficient central support is not available, we fall back to DBSCAN~\citep{ester1996density}, selecting the largest cluster and using its centroid to estimate depth. This fallback preserves robustness when the visible region is partial or noisy.

The final 3D picking point is obtained by combining the estimated depth with the image-space center of the target apple. Let $(x_c, y_c, z_c)$ denote the 3D point corresponding to the apple's image-space center. The picking position is computed as
\[
\frac{z_d}{z_c} \cdot (x_c, y_c, z_c),
\]
where $z_d$ is the estimated depth. This formulation is consistent with our previous system and supports stable downstream manipulation.

\paragraph{Localization Dataset and Evaluation.}
We evaluate localization using the occluded/unoccluded apple dataset introduced in~\cite{bhattacharya2025detect}. The dataset consists of paired observations with aligned RGB images and point clouds, where occlusion is defined per instance. Apples are detected independently in both views and matched using IoU, yielding 124 paired instances. Ground truth is defined from the unoccluded observation as the 3D point corresponding to the bounding box center, enabling direct comparison under different visibility conditions.

\begin{table}[t]
\centering
\caption{\textbf{Localization performance and efficiency.} 
Mean absolute error (MAE) and tail error (p95) are reported in millimeters for unoccluded and occluded samples. Latency is reported per instance using median (p50) and tail (p95) statistics. Center-weighted results under occlusion are reported only for instances with sufficient central support. The proposed method conditionally selects between center-weighted and DBSCAN localization, and its realized performance and latency depend on the distribution of visibility conditions in the dataset.}
\label{tab:localization_eval}
\setlength{\tabcolsep}{5pt}
\renewcommand{\arraystretch}{1.15}
\begin{tabular}{lcccccc}
\toprule
& \multicolumn{2}{c}{\textbf{Unoccluded (mm)}} 
& \multicolumn{2}{c}{\textbf{Occluded (mm)}} 
& \multicolumn{2}{c}{\textbf{Latency (ms)}} \\
\cmidrule(lr){2-3} \cmidrule(lr){4-5} \cmidrule(lr){6-7}
\textbf{Method} 
& \textbf{MAE} & \textbf{p95} 
& \textbf{MAE} & \textbf{p95} 
& \textbf{p50} & \textbf{p95} \\
\midrule
Center-weighted      & \textbf{1.054} & \textbf{2.960} & 6.876$^{\dagger}$ & 28.246$^{\dagger}$ & \textbf{0.130} & \textbf{0.149} \\
DBSCAN               & 5.183 & 10.600 & \textbf{8.104} & \textbf{19.750} & 10.36 & 29.20 \\
Proposed (Combined)  & \textbf{1.054} & \textbf{2.960} & 8.882 & 27.596 & 0.142 & 6.76 \\
\bottomrule
\end{tabular}
\\[-3pt]
\vspace{4pt}
{\footnotesize $^{\dagger}$ Evaluated only on occluded instances with sufficient central support.}
\end{table}

Table~\ref{tab:localization_eval} shows that center-weighted localization achieves the lowest error for unoccluded instances, and also performs well under occlusion when sufficient central support is available. In contrast, DBSCAN provides more consistent performance across general occluded cases. The proposed method combines both strategies, achieving the same accuracy as center-weighted localization in well-observed cases while remaining comparable to DBSCAN under limited visibility.

In practice, this conditional behavior enables efficient processing by applying the lightweight center-weighted estimator whenever sufficient support is available, while retaining DBSCAN as a robust fallback. As a result, the realized performance and latency of the combined method depend on the distribution of occlusion conditions in the dataset. This balance between accuracy, robustness, and efficiency allows the localization module to adapt to varying scene conditions and integrate effectively into the perception pipeline.

\subsection{Dual-Arm Harvest Coordination}\label{sec:coordination}
The dual-arm configuration requires a carefully designed coordination strategy to ensure efficient and reliable apple harvesting. Given the apple positions produced by the localization pipeline (Section~\ref{sec:perception}), the system maintains a tracked list of apples across frames, assigns each apple to one of the two arms, and sequences the per-arm targets in a way that avoids mutual interference and unnecessary fruit loss. Each arm then executes an independent harvest cycle using the single-arm motion stack detailed in Section~\ref{sec:motion_control} below. Three operational constraints govern the coordination:
\begin{enumerate}
\item The two arms must not attempt apple detachment simultaneously, because the shared vacuum source cannot deliver sufficient suction force to both end-effectors at the same time.
\item When a harvest attempt fails or attachment is lost, the affected arm must release its vacuum channel immediately to preserve suction capacity for the other arm.
\item A moving arm degrades perception quality by introducing motion blur and depth noise into the RGB-D frame; perception should therefore be triggered only when the requesting arm has cleared the camera's field of view.
\end{enumerate}

The localization pipeline produces an instantaneous list of apple centroids for each RGB-D frame without persistent identifiers, whereas coordination requires stable identities so that targets can be assigned, queued, and removed from consideration once picked. We therefore maintain a tracked apple list that is updated at every new perception result. Each new detection list is matched against the current set of tracked apples by solving a Hungarian assignment problem on the pairwise Euclidean distance matrix~\citep{kuhn1955hungarian}; matches within 10\,cm update the corresponding track, while unmatched detections initialize new tracks. A consolidation step periodically merges any pair of tracks lying within 5\,cm of each other to suppress duplicate identifiers caused by depth-channel jitter. Each track also stores a cumulative count of harvest attempts, which the dual-arm policy uses to remove apples that have already been picked or repeatedly missed from the candidate pool.

\begin{figure}[!htbp]
    \centering
    \includegraphics[width=0.75\linewidth]{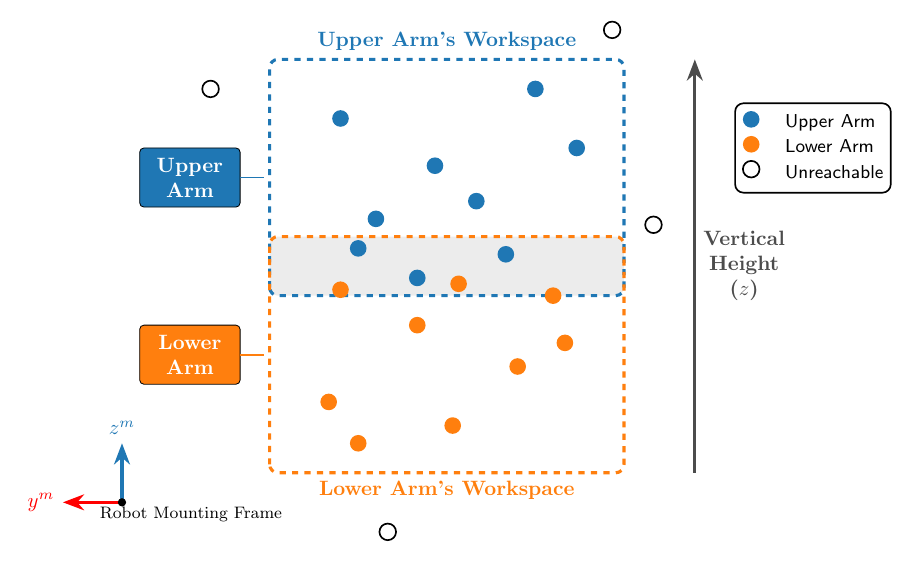}
    \caption{Apple assignment for the vertically-stacked dual-arm configuration (front view). The upper and lower arm workspaces overlap in the shaded region. Apples in the overlap zone are assigned to the vertically closer arm; apples outside both workspaces are marked as unreachable.}
    \label{fig:assignment}
\end{figure}

Building on this tracked apple list, the first step in each coordination cycle is to assign tracked apples to the two arms. Because the arms are stacked vertically with partially overlapping workspaces (Figure~\ref{fig:assignment}), each apple falls into one of three categories: reachable only by the upper arm, reachable only by the lower arm, or reachable by both. Apples outside both workspaces are discarded. For apples in the overlapping region, the system assigns each to the arm whose base is closer in the vertical ($z$) direction. Within each arm's assigned set, targets are sorted by depth so that apples closer to the canopy surface are harvested first, reducing the risk of the arm colliding with unharvested fruit deeper in the canopy. This assignment procedure follows a simplified version of the algorithm in our previous work~\citep{zhu2025advancement}, adapted from horizontal to vertical workspace partitioning.

Building on the coordination framework in our previous system~\citep{zhu2025advancement}, we synthesize a policy that enables the dual-arm system to operate continuously while strictly adhering to the constraints above. The policy is organized in three parts: the single-arm harvest cycle that defines the per-arm workflow (Section~\ref{sec:single_arm_cycle}), a temporal-logic formulation that encodes the dual-arm constraints and failure-recovery logic (Section~\ref{sec:vacuum_coord}), and an asynchronous vision-arm scheduling strategy that minimizes idle time between cycles (Section~\ref{sec:vision_arm_coord}).

\subsubsection{Single-Arm Harvest Cycle}\label{sec:single_arm_cycle}

\begin{figure}[!htbp]
    \centering
    \includegraphics[width=0.72\linewidth]{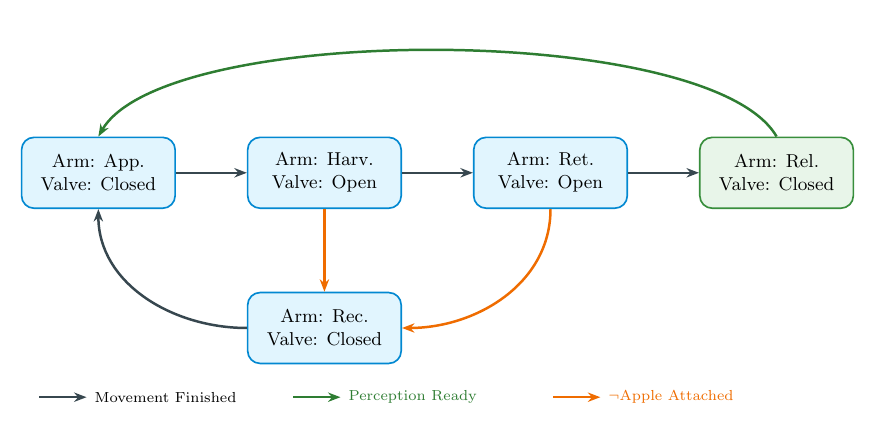}
    \caption{Single-arm harvest-cycle state machine. The arm progresses through \textit{Approaching}~$\to$~\textit{Harvesting}~$\to$~\textit{Retracting}~$\to$~\textit{Releasing} in the nominal case, or diverts to \textit{Recover} when attachment fails or is lost. This single-arm logic forms the building block for the dual-arm state graph in Figure~\ref{fig:state_machine}.}
    \label{fig:state_machine_single}
\end{figure}

Each arm operates in one of five states---\textit{Approaching}, \textit{Harvesting}, \textit{Retracting}, \textit{Releasing}, and \textit{Recover}---as illustrated in Figure~\ref{fig:state_machine_single}. The following paragraphs describe each state and the transitions between them.

\paragraph{Approaching.}
Given a target apple position $p_a = [x_a,y_a,z_a]^\top$ from the perception pipeline, the inverse kinematics (IK) module (presented below in Section~\ref{sec:kinematics}) first computes a \emph{buffer configuration} $q_b = [D_b,\theta_b,\varphi_b]^\top$ corresponding to a point $p_b$ located 10~cm in front of the apple along the approach direction. This buffer distance is large enough to absorb typical localization error while keeping the arm inside its reachable workspace. The arm is driven to $q_b$ using the jerk-continuous trajectory generator presented later in Section~\ref{sec:trajectory}, with the vacuum valve closed throughout the motion. The arm transitions to \textit{Harvesting} after reaching $q_b$ and satisfying the shared-vacuum scheduling condition (Section~\ref{sec:vacuum_coord}).

\begin{figure}[!htbp]
    \centering
    \includegraphics[width=0.85\linewidth]{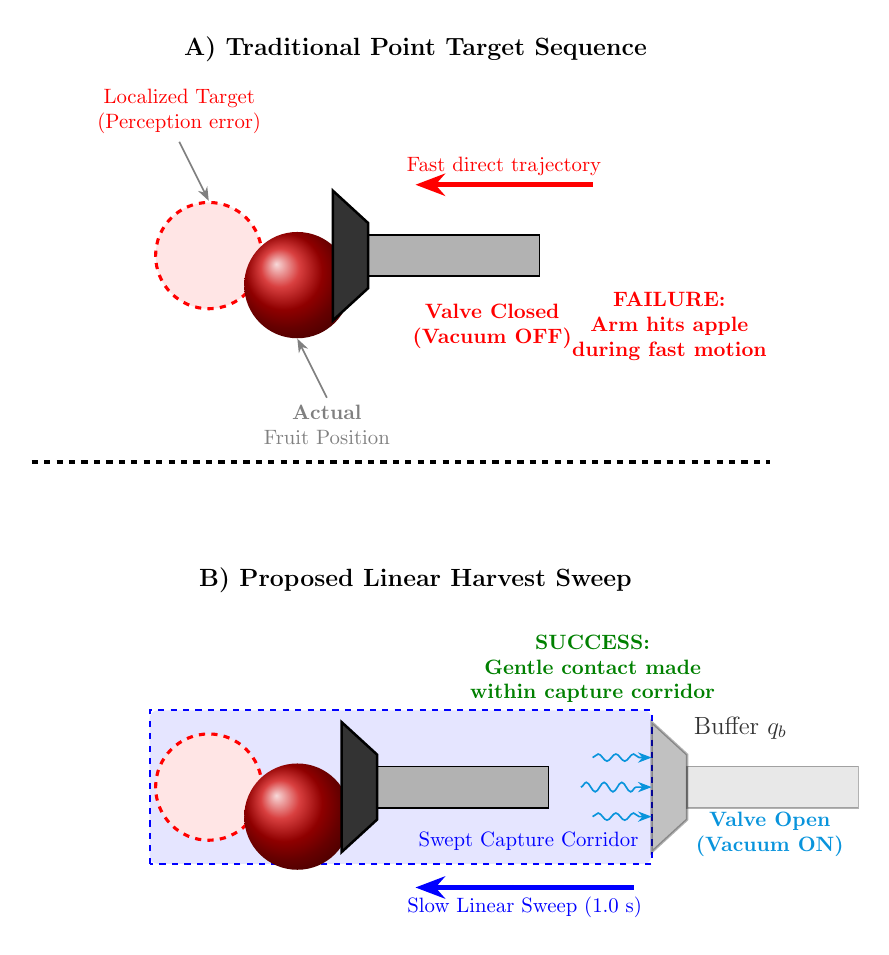}
    \caption{Comparison of harvest approach strategies under localization error. (A)~Traditional point-target approach: the arm moves directly to the estimated fruit position at full speed with the vacuum off; if the localized target is offset from the actual fruit, the end-effector may collide with the apple during fast approach, causing fruit damage or displacement. (B)~Proposed linear harvest sweep: the arm first reaches a buffer configuration $q_b$ via the jerk-continuous trajectory, then executes a slow 1.0\,s linear sweep toward the target with the vacuum on. The swept capture corridor absorbs the localization error, allowing suction attachment to occur anywhere along the corridor rather than at a single point.}
    \label{fig:linear_harvest_compare}
\end{figure}

\paragraph{Harvesting.}
Upon entering this state, the vacuum valve is opened, and the arm executes a slow linear sweep in joint space from the buffer configuration $q_b$ toward the target configuration $q_a$:
\begin{equation}\label{eq:linear_harvest}
q(t) = q_b + (q_a-q_b)\frac{t}{t_h}, \qquad t \in [0,\,t_h],
\end{equation}
where $t_h = 1.0$~s is the sweep duration. Although the remaining travel distance is short, the sweep is executed slowly by design: the duration is set by the time the vacuum needs to build up a reliable seal as the cup passes the fruit, not by the travel distance, and the low speed keeps contact with the apple gentle so that the fruit is not knocked loose before suction is established. Although this motion is linear in joint space, it generates a swept capture corridor in Cartesian space, allowing the end-effector to establish suction anywhere along the path rather than at a single precise contact point.
Compared with a traditional point-target approach that drives the arm directly to the estimated fruit position at full speed, this buffered linear sweep improves robustness by absorbing localization error and avoiding uncontrolled contact with the fruit during fast motion, as illustrated in Figure~\ref{fig:linear_harvest_compare}.

After suction is established---or when the prescribed sweep interval has elapsed---the end-effector executes a fixed $180^\circ$ rotation about joint $\rho$ to detach the fruit. This rotational detachment bends and breaks the stem while the apple remains attached to the suction cup, reducing dependence on pure pull force. Prior measurements on apple detachment found that the force required to detach fruit by pulling is significantly higher than the equivalent effort needed for twisting or rotation~\citep{lu2022design}, which motivates the use of rotational detachment here. Without this rotation, the arm would rely more heavily on translational pulling, which increases canopy disturbance, shakes neighboring apples, and raises the risk of both failed detachment and unwanted fruit drop. The arm subsequently transitions to \textit{Retracting} if attachment is confirmed by the pressure sensor; otherwise, it enters the \textit{Recover} state when attachment is not established or cannot be retained through the harvest interval.

\paragraph{Retracting.}
The arm returns toward the release side of the platform via a jerk-continuous trajectory while maintaining suction. If the fruit remains attached at the end of retraction, the arm proceeds to \textit{Releasing}. If pressure feedback indicates attachment loss during retraction, the valve is closed immediately to preserve vacuum for the other arm, and the arm diverts to \textit{Recover}.

\paragraph{Releasing.}
The end-effector is vented at the designated dropping location so that the apple falls onto the fruit handling module (Section~\ref{sec:fruit_handling}). The arm then waits for a new valid perception result before returning to \textit{Approaching}.

\paragraph{Recover.}
After a failed harvest attempt or mid-retraction attachment loss, the arm returns to a neutral configuration with the valve closed. Like \textit{Releasing}, the arm re-enters \textit{Approaching} once new perception data becomes available.
The following subsection formalizes this single-arm cycle together with the dual-arm coordination constraints using temporal logic.

\subsubsection{Temporal-Logic-Based Dual-Arm Policy}\label{sec:vacuum_coord}

Following our previous work~\citep{zhu2025advancement}, we employ Temporal Logic (TL)~\citep{baier2008principles} to specify the coordination policy that governs the dual-arm system. TL is a formal language for expressing goals and constraints over time. The core operators used in this work are $\neg$~(Not), $\land$~(And), $\lor$~(Or), $\rightarrow$~(Implies), $\bigcirc$~(Next), $\Diamond$~(Eventually), and $G$~(Always). For a detailed treatment and robotic applications, readers are referred to~\cite{baier2008principles,kress2009temporal}.

We define two sets of atomic propositions. \textbf{Environmental events}, denoted by $\phi_e = \{\mathit{Det}_i,\, \mathit{Att}_i\}$, model the external state: $\mathit{Det}_i$ is true when an apple has been detected and assigned to Arm~$i$, and $\mathit{Att}_i$ is true when an apple is currently attached to Arm~$i$'s end-effector, as inferred from the pressure sensor. \textbf{Robot actions}, denoted by $\phi_r = \{\mathit{App}_i,\, \mathit{Har}_i,\, \mathit{Ret}_i,\, \mathit{Rel}_i,\, \mathit{Rec}_i\}$, model the operational state: these correspond to the five harvest-cycle states described in Section~\ref{sec:single_arm_cycle}---\textit{Approaching}, \textit{Harvesting}, \textit{Retracting}, \textit{Releasing}, and \textit{Recover}. Valve control is implicit in each state: the valve opens upon entering \textit{Harvesting} and closes upon entering \textit{Recover} or \textit{Releasing}.

\paragraph{Workflow Specification.}
The single-arm harvest cycle defined in Section~\ref{sec:single_arm_cycle} and illustrated in Figure~\ref{fig:state_machine_single} is captured by the following temporal-logic formula. For each arm $i \in \{1,2\}$:
\begin{align}
\phi_{\text{workflow},i} =\;
&  G\big(\mathit{Det}_i \land (\mathit{Rel}_i \lor \mathit{Rec}_i) \rightarrow \bigcirc \mathit{App}_i\big) \nonumber\\
\land\; &  G\big(\bigcirc \mathit{Har}_i \rightarrow \mathit{App}_i\big) \nonumber\\
\land\; &  G\big(\mathit{Har}_i \land \mathit{Att}_i \rightarrow \bigcirc \mathit{Ret}_i\big) \nonumber\\
\land\; &  G\big(\mathit{Har}_i \land \neg\mathit{Att}_i \rightarrow \bigcirc \mathit{Rec}_i\big) \nonumber\\
\land\; &  G\big(\mathit{Ret}_i \land \mathit{Att}_i \rightarrow \bigcirc \mathit{Rel}_i\big) \nonumber\\
\land\; &  G\big(\mathit{Ret}_i \land \neg\mathit{Att}_i \rightarrow \bigcirc \mathit{Rec}_i\big).
\label{eq:workflow_ltl}
\end{align}
The first clause states that a new harvest cycle begins only when an apple has been detected and assigned ($\mathit{Det}_i$) and the arm has completed its previous cycle by reaching \textit{Releasing} or \textit{Recover}. The second clause is a necessary condition: the arm may enter \textit{Harvesting} only after \textit{Approaching} has completed; however, completing \textit{Approaching} does not guarantee an immediate transition, as the arm may need to wait until the shared-vacuum constraint (below) is satisfied. The third and fourth clauses encode the branching after \textit{Harvesting}: if attachment is confirmed ($\mathit{Att}_i$), the arm proceeds to \textit{Retracting}; otherwise, it immediately transitions to \textit{Recover}, closing the valve to preserve vacuum capacity for the other arm. The fifth and sixth clauses apply the same branching during \textit{Retracting} to handle mid-retraction attachment loss. Since both arms follow the same structure, the overall workflow constraint is $\phi_{\text{workflow}} = \phi_{\text{workflow},1} \land \phi_{\text{workflow},2}$.

\begin{figure}[!htbp]
    \centering
    \includegraphics[width=\linewidth]{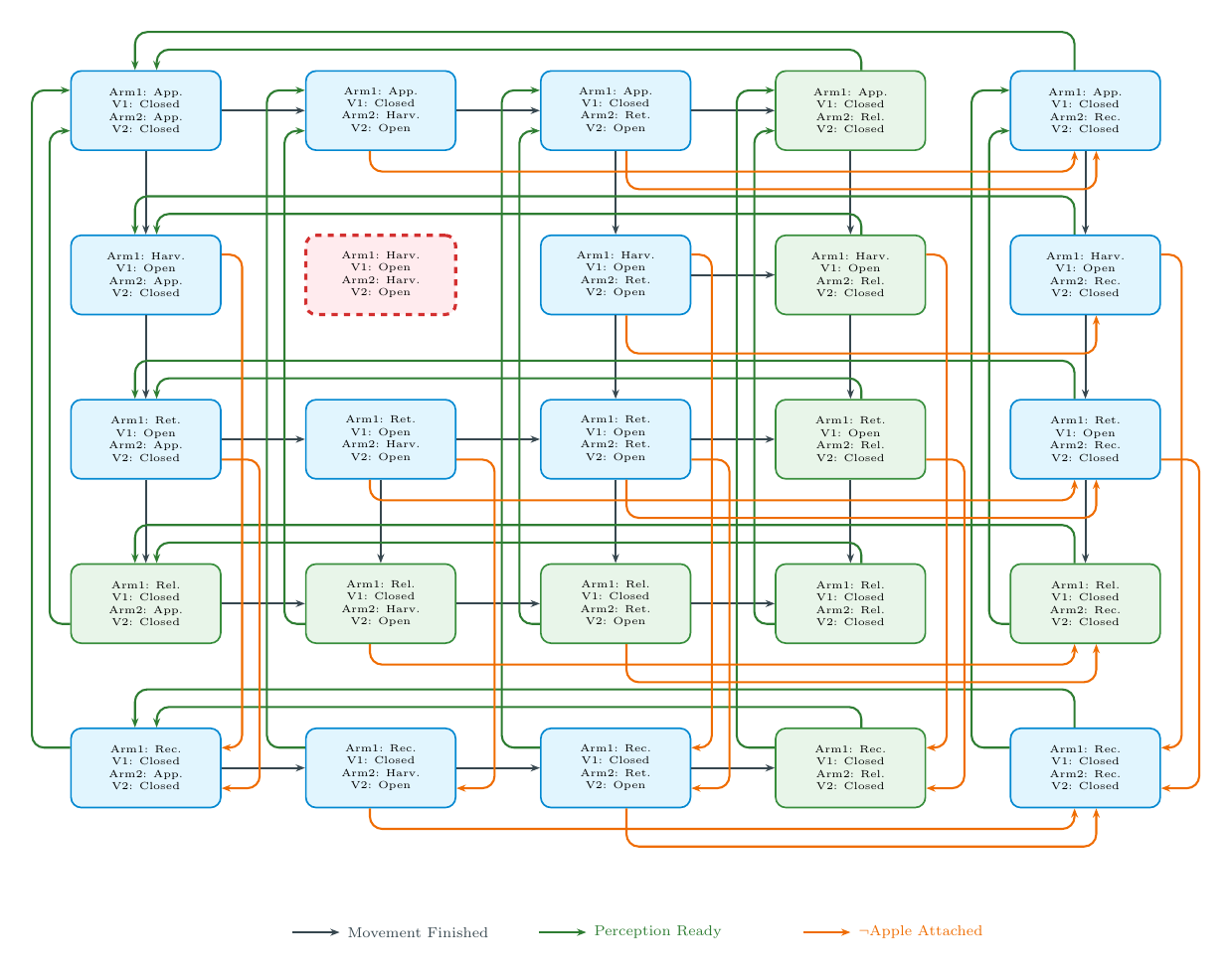}
    \caption{Dual-arm coordination state graph derived from the combined specification $\phi$ (Eq.~\ref{eq:combined_ltl}). Each node denotes a pair of arm states and valve configurations. Black arrows represent nominal progression after motion completion, green arrows denote transitions enabled by newly valid perception results, and orange arrows denote failure-triggered transitions when attachment is not retained.}
    \label{fig:state_machine}
\end{figure}

\paragraph{Shared-Vacuum Constraint.}
The two arms share a single vacuum source. If both arms attempt apple attachment simultaneously, the suction force available to each arm is reduced, degrading detachment reliability. We therefore require that at most one arm may be in the \textit{Harvesting} state at a time:
\begin{equation}\label{eq:harvest_mutex}
\phi_{\text{vacuum}} = G\neg(\mathit{Har}_1 \land \mathit{Har}_2).
\end{equation}
An important exception improves efficiency without violating this constraint: once one arm has secured an apple and entered \textit{Retracting}, its suction line is effectively sealed by the attached fruit. The second arm is then permitted to begin \textit{Harvesting}, because the sealing ensures that vacuum is not split between two open channels. In practice, this allows useful overlap between one arm retracting with fruit and the other arm starting its next harvest attempt.

\paragraph{Goal Condition.}
The coordination policy must ensure that the system continues harvesting as long as apples remain available. This liveness requirement is expressed as
\begin{equation}\label{eq:goal_ltl}
\phi_{\text{goal}} = G(\mathit{Det}_1 \rightarrow \Diamond\,\mathit{Rel}_1) \land G(\mathit{Det}_2 \rightarrow \Diamond\,\mathit{Rel}_2).
\end{equation}
That is, whenever an apple is detected and assigned to arm~$i$, the arm should eventually reach the \textit{Releasing} state, completing a full harvest cycle.

\paragraph{Combined Specification.}
The overall dual-arm coordination policy is the conjunction of the workflow, mutual-exclusion, and goal specifications:
\begin{equation}\label{eq:combined_ltl}
\phi = \phi_{\text{workflow}} \land \phi_{\text{vacuum}} \land \phi_{\text{goal}}.
\end{equation}
This combined specification is realized as the directed state graph in Figure~\ref{fig:state_machine}. Each node represents a pair of arm states together with the valve configuration; the edges encode exactly the transitions permitted by $\phi$. The graph serves as the runtime coordination controller: starting from the initial node, the system determines the next action for each arm based on the current state pair, sensor feedback, and the constraint set above.

\subsubsection{Asynchronous Vision-Arm Scheduling}\label{sec:vision_arm_coord}
During harvesting, the arms regularly traverse the camera's field of view. Analysis of data from our previous field season~\citep{zhu2025advancement} revealed that a moving arm in the scene degrades perception quality in three ways. First, the moving arm introduces motion blur in the RGB image. Second, and more critically, it degrades depth measurements: the ToF sensor produces noisy point clouds near a moving arm, corrupting the localization of nearby apples. Third, under highly dynamic conditions, the RGB and depth channels exhibit synchronization offsets, causing misalignment between the color image and the corresponding depth map. In addition, the arm can directly occlude target apples. To improve perception reliability, we adopt a strategy that combines release-position sampling with independent acceptance logic.

Each harvest cycle ends with the arm moving toward a predefined release position on the non-picking side of the platform. Once the arm crosses into the non-occluding region, a new RGB-D frame is acquired and processed, ensuring that every perception frame used for localization is captured with both arms clear of the workspace and free of the motion-blur and depth-noise artifacts described above. The resulting perception result is then validated \emph{independently} for each arm. Because the two arms are vertically stacked, their target regions are largely separated: in the vast majority of scenarios, Arm~1 does not enter the region assigned to Arm~2, and vice versa. Therefore, when deciding whether to accept a localization result for Arm~$i$, the system only requires Arm~$i$ itself to have cleared the camera's field of view; it does not wait for the other arm to finish its cycle. Compared with a na\"ive strategy that waits until both \textit{Releasing} and \textit{Recover} fully complete before triggering a new perception pass, this decoupled early-acceptance mechanism prevents one arm from idling unnecessarily while the other is still retracting or releasing, reclaiming approximately 1~s of otherwise idle time per cycle. The corresponding perception-ready event is represented by the green transitions in Figure~\ref{fig:state_machine}, which return an arm from \textit{Releasing} or \textit{Recover} to \textit{Approaching} as soon as its own perception data become usable.

\subsubsection{Indoor Coordination Validation}\label{subsec:indoor}

Before deploying the system to the field, we validated the coordination strategy through controlled indoor experiments comparing three modes:
\begin{itemize}
    \item \textit{Serialized} (lower bound): a single vacuum pump with strictly sequential arm operation, representing a system without coordination.
    \item \textit{Full Parallel} (upper bound): two independent vacuum pumps allowing both arms to operate with no scheduling constraints, representing the theoretical performance ceiling.
    \item \textit{Proposed}: our method with a single shared vacuum pump, implementing vision--arm asynchrony and mutex-based harvesting scheduling.
\end{itemize}

Nine experimental groups were conducted, each involving 11 to~20 simulated apple positions. The purpose of these indoor experiments is to isolate the efficiency of the coordination layer---vacuum arbitration and vision--arm asynchronous scheduling---rather than to evaluate picking success. The per-attempt throughput is governed by the arm motion and scheduling timing, each arm executes its full harvest cycle (including the vacuum sweep, rotational detachment, and retraction motions) regardless of whether a physical fruit is present, so the measured cycle timing is independent of fruit attachment. Using fixed simulated apple positions further ensures that the three coordination modes are compared against identical, repeatable target sets, avoiding the confounding effects of perception error, attachment variability, and canopy differences that arise in the field. Accordingly, vacuum attachment success is not used as an indoor metric; instead, we measure the wall-clock elapsed time to complete all $N$ arm cycles and derive the system throughput as the elapsed time per attempt (seconds per attempt). Real-fruit attachment success and field cycle times are characterized separately in the field trials (Section~\ref{subsec:field}).

Table~\ref{tab:indoor} summarizes the results across all nine groups. The proposed method achieves a mean throughput of $3.65 \pm 0.16$~s/attempt, compared to $3.52 \pm 0.22$~s/attempt for the unconstrained parallel operation and $6.44 \pm 0.08$~s/attempt for serialized cycling. This represents a $1.76\times$ speedup over the serialized operation, while incurring only a $3.7\%$ overhead relative to the two-pump upper bound. Compared to our previous dual-arm system~\citep{zhu2025advancement}, which achieved a system throughput of 5.29~s/attempt under equivalent indoor conditions, the proposed coordination strategy reduces per-attempt processing time by 31\%. These results demonstrate the effectiveness of vision--arm asynchrony in maximizing utilization of a single vacuum source.

\begin{table}[htbp]
    \centering
    \caption{Indoor coordination experiment results. Throughput is reported in seconds per attempt (elapsed time / attempt count). Lower is better.}
    \label{tab:indoor}
    \begin{tabular}{crrr}
        \hline
        Group ($N$) & \textit{Proposed} & \textit{Full Parallel} & \textit{Serialised} \\
        \hline
        12 & 3.89 & 3.86 & 6.62 \\
        16 & 3.68 & 3.62 & 6.42 \\
        20 & 3.58 & 3.27 & 6.43 \\
        14 & 3.46 & 3.22 & 6.35 \\
        18 & 3.82 & 3.65 & 6.48 \\
        20 & 3.75 & 3.57 & 6.45 \\
        18 & 3.45 & 3.27 & 6.40 \\
        11 & 3.52 & 3.51 & 6.38 \\
        16 & 3.68 & 3.67 & 6.43 \\
        \hline
        Mean $\pm$ Std & $3.65 \pm 0.16$ & $3.52 \pm 0.22$ & $6.44 \pm 0.08$ \\
        \hline
    \end{tabular}
\end{table}

\subsection{Manipulator Low-Level Control}\label{sec:motion_control}
This subsection details the single-arm motion stack invoked at each step of the harvest cycle defined in Section~\ref{sec:coordination}. Once a target apple has been localized in the camera frame (Section~\ref{sec:perception}), the manipulator must convert that position into joint commands, generate a smooth trajectory, and track it in real time. The motion pipeline comprises three layers: kinematic analysis (Section~\ref{sec:kinematics}), jerk-continuous trajectory generation and tracking (Section~\ref{sec:trajectory}), and a safety filter based on Control Barrier Functions (Section~\ref{sec:cbf}).

\subsubsection{Kinematic Analysis}\label{sec:kinematics}

The manipulator retains the 4-DOF kinematic structure of our earlier single-arm and dual-arm harvesters; the kinematic model and the closed-form inverse-kinematics solution summarized below were established in our prior work~\citep{zhang2024automated, zhu2025advancement} and are restated here to keep the paper self-contained.

\begin{figure}[!htbp]
    \centering
    \includegraphics[width=0.65\linewidth]{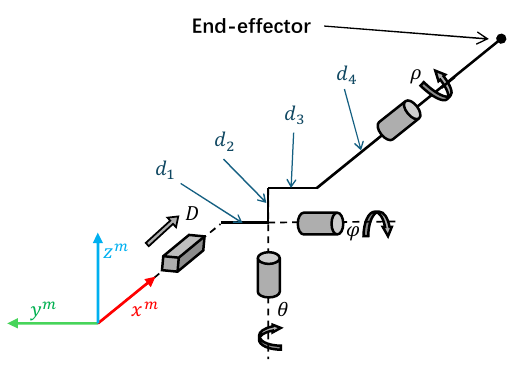}
    \caption{Kinematic model of a single 4-DOF manipulator. The mounting frame $\{x^m,y^m,z^m\}$ is fixed to the robot chassis. Joint $\theta$ rotates the arm assembly about the vertical axis; $D$ extends the arm along the prismatic stage; $\varphi$ tilts the outer link; and $\rho$ rotates the end-effector for fruit detachment. The link parameters $d_1$--$d_4$ define the geometric offsets between successive joints.}
    \label{fig:arm_configuration}
\end{figure}

Each manipulator has four degrees of freedom: one prismatic joint $D$ for linear extension, two revolute joints $(\theta, \varphi)$ that determine the approach direction, and one end-effector rotation joint $\rho$ used exclusively for fruit detachment after suction engagement (see Figure~\ref{fig:arm_configuration}). Because $\rho$ does not affect the end-effector position, the kinematic model for approach and retraction planning involves only the three positioning joints $q = [D, \theta, \varphi]^\top$. The system parameters are listed in Table~\ref{tab:arm_parameters}.

Let $(x^m,y^m,z^m)$ denote the mounting frame of the manipulator (Figure~\ref{fig:arm_configuration}). The end-effector position $p_e = [x_e,\, y_e,\, z_e]^\top$ in this frame is given by
\begin{equation}\label{eq:fk}
p_e = \begin{bmatrix}
d_4 \cos\varphi\cos\theta + d_3 \sin\theta + D \\[2pt]
-d_4 \cos\varphi\sin\theta + d_3 \cos\theta + d_1 \\[2pt]
-d_4 \sin\varphi + d_2
\end{bmatrix},
\end{equation}
where $d_1$ and $d_2$ are fixed offsets from the mounting origin to the first joint axis, $d_3$ is a lateral offset at the elbow, and $d_4$ is the length of the outer link from the $\varphi$-joint to the end-effector tip. Given a target apple position $p_a = [x_a,\, y_a,\, z_a]^\top$ in the mounting frame, the decoupled structure of~\eqref{eq:fk} admits a closed-form inverse kinematic solution:
\begin{equation}\label{eq:ik}
\left\{
\begin{array}{l}
\varphi_d = \sin^{-1}\!\!\left(\dfrac{d_2 - z_a}{d_4}\right), \\[8pt]
\big(d_4^2\cos^2\!\varphi_d + d_3^2\big)\sin^2\!\theta_d
+ 2(y_a - d_1)\,d_4\cos\varphi_d\;\sin\theta_d
+ (y_a - d_1)^2 - d_3^2 = 0, \\[6pt]
D_d = x_a - d_4\cos\varphi_d\cos\theta_d - d_3\sin\theta_d,
\end{array}
\right.
\end{equation}

\begin{table}[!htbp]
\centering
\caption{Kinematic parameters of the two manipulators. All length parameters are in meters; angular limits are in radians. Minor differences in $d_3$ and $d_4$ between the two arms reflect manufacturing and assembly tolerances.}
\label{tab:arm_parameters}
\begin{tabular}{lcc}
\toprule
Parameter & Arm~1 & Arm~2 \\
\midrule
$d_1$ (m) & $-0.150$ & $-0.150$ \\
$d_2$ (m) & $\phantom{-}0.079$ & $\phantom{-}0.079$ \\
$d_3$ (m) & $-0.116$ & $-0.120$ \\
$d_4$ (m) & $\phantom{-}0.871$ & $\phantom{-}0.873$ \\
\midrule
$D$ range (m) & $[-0.01,\;0.56]$ & $[-0.01,\;0.56]$ \\
$\theta$ range ($\mathrm{rad}$) & $[-0.52,\;0.40]$ & $[-0.52,\;0.40]$ \\
$\varphi$ range ($\mathrm{rad}$) & $[-0.52,\;0.30]$ & $[-0.52,\;0.35]$ \\
\bottomrule
\end{tabular}
\end{table}

where $\theta_d$ is obtained by taking the smaller root of the quadratic in $\sin\theta_d$.

\paragraph{Coordinate Transformation and Workspace Check.}
The perception pipeline (Section~\ref{sec:perception}) returns apple positions in the camera frame. A pre-calibrated rigid-body transformation $T^m_c$ converts each position into the corresponding mounting frame before the inverse kinematics is applied. After computing $q_d = [D_{d}, \theta_{d}, \varphi_{d}]^\top$ via~\eqref{eq:ik}, the system verifies that the solution lies within the admissible joint range. Because the prismatic stage and the arm body can collide at certain joint combinations, the feasible joint space is defined as the union of two rectangular regions that were determined experimentally; an apple whose IK solution falls outside both regions is marked as unreachable and excluded from the current harvest cycle.

\subsubsection{Jerk-Continuous Trajectory Generation and Tracking}\label{sec:trajectory}
In our previous dual-arm system~\citep{zhu2025advancement}, approach and return motions were generated by quintic splines. Although quintic trajectories provide continuous position, velocity, and acceleration, they can still excite visible vibration in the linear stage when executed aggressively. To reduce this effect, we use a 7th-order polynomial for each of the major approach joints $i \in \{D, \theta, \varphi\}$:
\begin{equation}\label{eq:7th_order_poly}
q_{i,r}(t) = a_7 t^7 + a_6 t^6 + a_5 t^5 + a_4 t^4 + a_3 t^3 + a_2 t^2 + a_1 t + a_0,
\end{equation}
where $t \in [0, t_{\text{total}}]$ is time and $\{a_0,\ldots,a_7\}$ are the polynomial coefficients.

The coefficients are determined from eight boundary conditions:
\begin{align}
q_i(0) &= q_{i,c}, &
\dot{q}_i(0) &= \dot{q}_{i,c}, &
q_i^{(2)}(0) &= q_{i,c}^{(2)}, &
q_i^{(3)}(0) &= J_i \, \mathrm{sgn}(q_{i,d}-q_{i,c}), \label{eq:bc_init}\\
q_i(t_{\text{total}}) &= q_{i,d}, &
\dot{q}_i(t_{\text{total}}) &= 0, &
q_i^{(2)}(t_{\text{total}}) &= 0, &
q_i^{(3)}(t_{\text{total}}) &= 0. \label{eq:bc_terminal}
\end{align}
Here $\dot{q}_i$ is the joint velocity, and $q_i^{(2)}$ and $q_i^{(3)}$ denote the second and third time derivatives (acceleration and jerk); $q_{i,c}$ and $q_{i,d}$ denote the current and desired joint positions. The injected initial jerk term gives the motion a smooth but decisive start, with empirically selected magnitudes $J_\theta = J_\varphi = 10$~rad/s$^3$ and $J_D = 2$~m/s$^3$.

To synchronize all major joints, the total motion time is computed from the largest relative displacement:
\begin{equation}\label{eq:sync_time}
t_{\text{total}} = \max \left(
\max\left[
\frac{|\Delta D|}{D_{\text{lim}}},
\frac{|\Delta \theta|}{\theta_{\text{lim}}},
\frac{|\Delta \varphi|}{\varphi_{\text{lim}}}
\right] t_{\max},
t_{\min}
\right),
\end{equation}
where $t_{\max}$ and $t_{\min}$ are tunable parameters that govern the motion aggressiveness of the arm, and $\{D_{\text{lim}}, \theta_{\text{lim}}, \varphi_{\text{lim}} \}$ denote the allowable motion ranges of respective joints. The resulting jerk-bounded trajectories are designed to suppress linear-stage oscillation, thereby shortening the effective settling time before harvesting and release.

\paragraph{Trajectory Tracking.}
The reference trajectory is tracked by a joint-space proportional controller with feedforward compensation. For each positioning joint $i \in \{D,\theta,\varphi\}$, the nominal velocity command is
\begin{equation}\label{eq:nominal_command}
u_i = \dot{q}_{i,d}(t) + k_i\big(q_{i,d}(t) - q_{i,c}\big),
\end{equation}
where $\dot{q}_{i,d}(t)$ and $q_{i,d}(t)$ are the desired velocity and position from the reference polynomial, $q_{i,c}$ is the measured joint position, and $k_i > 0$ is a proportional gain. The feedforward term $\dot{q}_{i,d}$ drives the joint along the planned profile, while the proportional term corrects deviations caused by communication latency and timing synchronization between the trajectory planner and the servo drives, which maintain their own internal PID (proportional--integral--derivative) loops.

Compared to the Cartesian-space nonlinear tracking controller used in our previous system~\citep{zhu2025advancement}, this joint-space formulation offers two practical advantages. First, it avoids the Jacobian inversion required by Cartesian controllers, which may be ill-conditioned under certain configurations. Second, indoor comparison tests showed improved settling behavior at the end of each trajectory: because the 7th-order polynomial already guarantees zero terminal velocity, acceleration, and jerk, the joint-space controller smoothly converges to the target configuration without the residual oscillation occasionally observed with the previous Cartesian controller near singular configurations. The nominal command~\eqref{eq:nominal_command} is then passed through the CBF safety filter described next.

\subsubsection{General Joint-Limit Safety Filter via CBF}\label{sec:cbf}
Trajectory smoothness alone is insufficient for reliable operation; joint commands must also remain inside the admissible configuration space. We therefore apply a Control Barrier Function (CBF) filter~\citep{ames2016control} at the velocity-command stage to enforce joint limits for the three positioning joints $q = [D,\theta,\varphi]^\top$. The end-effector rotation joint $\rho$ is excluded, as it does not affect the end-effector position and is not subject to a range limit.

For each joint $i$, we define lower and upper barrier functions
\begin{equation}
h^{-}_{i}(q) = q_i - q_{i,\min}, \qquad
h^{+}_{i}(q) = q_{i,\max} - q_i,
\end{equation}
so that admissible motion corresponds to $h^{-}_{i}(q) \ge 0$ and $h^{+}_{i}(q) \ge 0$ for all joints. Given the nominal command $u$ from~\eqref{eq:nominal_command}, the filtered command $\dot{q}^{*}$ is obtained by solving the quadratic program (QP)
\begin{align}
\dot{q}^{*} = \arg\min_{\dot{q}} \; &\frac{1}{2}\|\dot{q}-u\|^2 \label{eq:cbf_qp}\\
\text{subject to} \; &
\nabla h^{-}_{i}(q)^\top \dot{q} \ge -\gamma h^{-}_{i}(q), \nonumber\\
&
\nabla h^{+}_{i}(q)^\top \dot{q} \ge -\gamma h^{+}_{i}(q), \qquad  i \in \{D, \theta, \varphi\}, \nonumber
\end{align}
with $\gamma = 5$ in our implementation. Because each barrier function depends on only a single joint variable, the gradient vectors $\nabla h^{\pm}_i$ are axis-aligned and the QP decouples into per-joint scalar problems that can be solved in closed form. The filter is implemented in the motor communication layer and executes at the servo loop rate with negligible computational overhead.


\section{Experiments}\label{sec:experiment}

\subsection{Field Trial Results}\label{subsec:field}

The system was tested and evaluated at commercial orchards during the 2025 harvest season in both Michigan and Washington State, USA. Preliminary testing was conducted in Michigan from August through October~2025, where fruit-quality grading experiments were also performed. The primary field experiment was carried out in Washington State in four sessions in October~2025, and the harvesting results reported below are drawn from these sessions.

\begin{figure}[htbp]
	\centering
	\begin{subfigure}[b]{0.3\textwidth}
		\centering
		\includegraphics[width=1\textwidth]{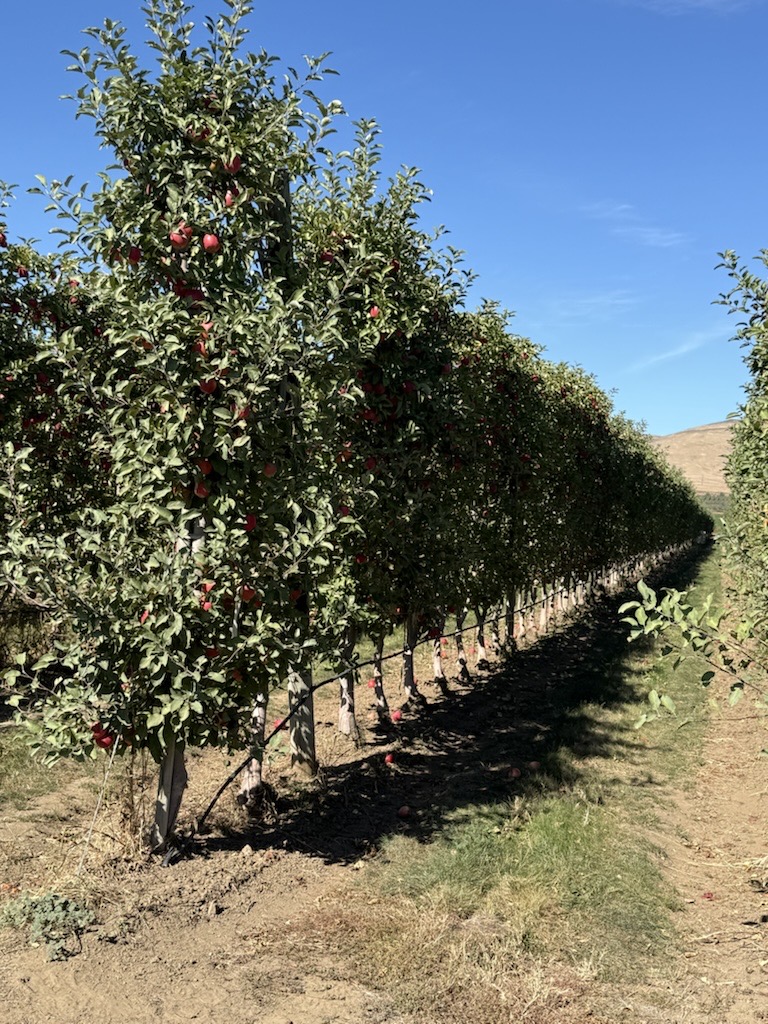}
		\caption{}
		\label{fig:jeff}
	\end{subfigure}
	\hspace{15pt}
	\begin{subfigure}[b]{0.532\textwidth}
		\centering
		\includegraphics[width=1\textwidth]{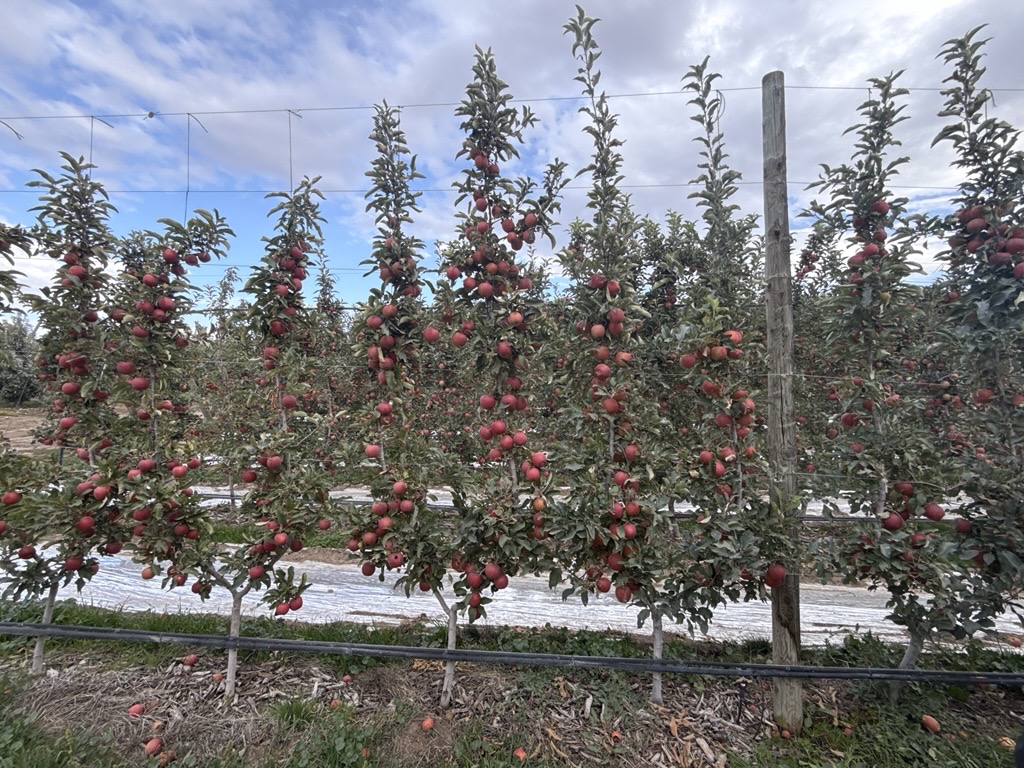}
		\caption{}
		\label{fig:keith}
	\end{subfigure}
    \caption{Field trial sites: (a)~Nicoter apple orchard with a vertical fruiting wall structure (Wenatchee, WA) and (b)~Envy apple orchard with a tall spindle structure (Yakima, WA).}
    \label{fig:orchards}
\end{figure}

The Washington trials were conducted at two commercial orchards (Figure~\ref{fig:orchards}). The first site, in Wenatchee, WA, had Nicoter apple trees trained to a vertical fruiting wall structure; the second site, in Yakima, WA, had Envy apple trees grown in a tall spindle configuration. The robot was mounted on a trailer, and operation followed a stop-and-go procedure: the tractor halted at each harvesting location, the operator made minor platform adjustments to align the robot's workspace with the target tree, and autonomous harvesting then proceeded until all reachable apples in the workspace had been attempted or assigned. Compared to our previous horizontal configuration~\citep{zhu2025advancement}, which required repeated lateral and vertical repositioning to cover multiple adjacent trees, the vertical layout of the current system allows each stop to target a single tree with only a small forward-backward translation and a vertical fine-adjustment of the platform, substantially reducing positioning time. During field trials, the arm movement speed was set to 95\% of the maximum actuation capacity by configuring $t_{\max}=1.5$~s and $t_{\min}=1.0$~s in~\eqref{eq:sync_time}; the remaining margin was retained to limit disturbance to the orchard canopy.

Across the four Washington sessions, a total of approximately 250 trees were tested. From these, 195 harvesting positions were selected for performance evaluation, with the remaining positions excluded due to ongoing system calibration or parameter tuning during those sessions. Table~\ref{tab:field} summarizes harvesting performance across all four sessions: the 1,738 selected arm cycles yielded an overall success rate of 80.0\% (79.7\% for Arm~1, 80.4\% for Arm~2), with a mean per-arm cycle time of $7.53 \pm 0.95$~s at 95\% actuation capacity ($n = 916$ cycles).

\begin{table}[htbp]
    \centering
    \caption{Field trial results by session. Positions denote the number of harvesting stops selected for evaluation. Attempts are the combined arm-cycle total of both arms. SR~=~success rate. Cycle time is the mean per-arm cycle time at 95\% actuation capacity.}
    \label{tab:field}
    \begin{tabular}{lllrrrrr}
        \hline
        Date & Variety & Structure & Positions & Attempts & SR (Arm~1) & SR (Arm~2) & Cycle Time \\
        \hline
        Oct.\ 8  & Nicoter & Fruiting wall & 57  & 438  & 78.4\% & 80.4\% & 7.85~s \\
        Oct.\ 9  & Nicoter & Fruiting wall & 109 & 870  & 79.7\% & 82.2\% & 7.62~s \\
        Oct.\ 10 & Envy    & Tall spindle  & 15  & 197  & 79.1\% & 80.2\% & 7.38~s \\
        Oct.\ 13 & Envy    & Tall spindle  & 14  & 233  & 83.0\% & 75.6\% & 7.45~s \\
        \hline
        Overall  & --      & --            & 195 & 1,738 & 79.7\% & 80.4\% & 7.53~s \\
        \hline
    \end{tabular}
\end{table}

The success rate reported here is computed on a per-attempt basis: each arm cycle is counted independently as a success or failure, regardless of whether the same apple was targeted in a prior attempt.
This differs from the per-apple metric used in~\citet{zhu2025advancement}, where an apple eventually harvested after multiple attempts was recorded as a single success; that system reported 80.7\% and 79.7\% for the two arms.
Because the per-attempt metric does not credit retries, the 80.0\% success rate reported here represents a conservative lower bound on the effective per-apple harvesting rate.
The mean per-arm cycle time of 7.53~s represents a 9.2\% reduction from the 8.29~$\pm$~1.02~s reported in~\citet{zhu2025advancement}, attributable primarily to the jerk-continuous trajectory controller (Section~\ref{sec:trajectory}), which reduces mid-motion vibration and enables faster execution.
We note that the vertical arm configuration requires a longer travel path between the harvest and release positions compared to the horizontal layout in~\citet{zhu2025advancement}; this geometric trade-off is a deliberate design choice and is more than offset by the trajectory improvement.

Observed failure modes fall into three categories.
\textit{Perception failures} occurred when heavy leaf occlusion reduced the visible apple surface, causing the localization algorithm to miss or mislocate the fruit.
\textit{Vacuum failures} arose either when the end-effector reached the target position but could not form a seal, or when suction was established but insufficient to detach the fruit; the rotational detachment strategy (Section~\ref{sec:single_arm_cycle}) mitigates the latter but does not eliminate it entirely.
\textit{Environmental interference} included leaves obstructing the suction cup, branches struck during approach causing the apple to shift before contact, and branches contacted during retraction dislodging an already-attached fruit.
These failure modes are largely consistent with those reported in~\citet{zhu2025advancement}; the primary improvement in the current system is the elimination of camera exposure failures that had previously degraded outdoor perception.

\subsection{Fruit Damage Assessment}\label{subsec:bruise}

Fruit quality was evaluated through two complementary assessments: an immediate USDA grading study on fruit harvested at Michigan orchards and a delayed damage inspection on fruit from the Washington field trials after one month of cold storage. Both studies use two shared baseline conditions, \textit{Platform}---fruit were hand-picked and then passed through the fruit-handling and bin-filling system---and \textit{Robotic harvest}---fruit underwent the complete robotic picking and handling process; the Washington study additionally includes a \textit{Hand} condition in which fruit were hand-picked and evaluated directly, providing a baseline free of mechanical handling.

\textbf{Immediate USDA grading.}
Following the methodology introduced in~\citet{zhang2024automated}, EverCrisp and Pink Lady apples harvested at Michigan orchards were graded under USDA categories: Extra Fancy (grades~1--4), Fancy, and Downgraded; Table~\ref{tab:bruise} summarises the resulting grade distributions.

\begin{table}[htbp]
    \centering
    \caption{Immediate USDA grade distribution for Michigan-harvested EverCrisp and Pink Lady fruit.}
    \label{tab:bruise}
    \begin{tabular}{llrrrr}
        \hline
        Condition & Variety & $N$ & Extra Fancy & Fancy & Downgraded \\
        \hline
        Platform        & EverCrisp & 150 & 97.3\% & 2.7\%  & 0      \\
        Platform        & Pink Lady & 149 & 98.6\% & 1.3\%  & 0      \\
        Robotic harvest & EverCrisp & 147 & 98.0\% & 2.0\%  & 0      \\
        Robotic harvest & Pink Lady & 150 & 84.7\% & 14.0\% & 1.3\%  \\
        \hline
    \end{tabular}
\end{table}

Across the robotically harvested fruit, 91.2\% (271/297) remained in the Extra Fancy category, 8.1\% (24/297) were graded as Fancy, and only 0.7\% (2/297) were downgraded.
For the same cultivars in the platform baseline, 98.0\% (293/299) remained Extra Fancy and no fruit were downgraded.
The outcome was cultivar dependent: robotically harvested EverCrisp remained comparable to the platform baseline (98.0\% versus 97.3\% Extra Fancy), whereas Pink Lady exhibited a lower Extra Fancy fraction of 84.7\% and accounted for the only downgraded fruit in the robotic condition. Notably, the Pink Lady platform baseline was the highest of all conditions (98.6\% Extra Fancy), so this reduction appears to arise specifically from the robotic picking step rather than from fruit handling, likely because Pink Lady is more sensitive to the suction-and-rotation detachment forces than EverCrisp.

\textbf{Delayed damage inspection.}
Nicoter fruit harvested on October~9 during the Washington trials were inspected after one month of cold storage to assess delayed damage. The robotic condition was tested at two actuation capacity levels (60\% and 95\%), comprising six replicates (three per level), while the hand and platform conditions each had three replicates. Table~\ref{tab:delayed_bruise} summarizes the resulting damage rates.

\begin{table}[htbp]
    \centering
    \caption{Delayed fruit-quality inspection on Nicoter fruit harvested on October~9 and inspected after one month of cold storage. Damage categories: Bruise~=~visible bruise marks; Stemless~=~stem detached during picking (not applicable to hand-picked conditions); Punctures~=~skin puncture wounds; Limb~Rub~=~surface abrasion from branch contact. Values are percentages of total fruit.}
    \label{tab:delayed_bruise}
    \begin{tabular}{llrrrrrr}
        \hline
        Condition & Capacity & Reps & $N$ & Bruise & Stemless & Punctures & Limb Rub \\
        \hline
        Robotic harvest & 60\% & 3 & 188 & 4.9\% & 34.0\% & 19.1\% & 1.3\% \\
        Robotic harvest & 95\% & 3 & 198 & 2.4\% & 33.9\% & 11.8\% & 0.7\% \\
        Hand            & --   & 3 & 173 & 1.3\% & --     &  6.5\% & 0     \\
        Platform        & --   & 3 & 188 & 1.1\% & --     & 10.4\% & 1.1\% \\
        \hline
    \end{tabular}
\end{table}

Bruise rates were low across all conditions, with the robotic harvest at 2.4--4.9\% compared with 1.1--1.3\% for the non-robotic baselines.
Puncture wounds were observed in all conditions, including the hand-picked baseline (6.5\%), suggesting that a portion of puncture damage originates from pre-existing field conditions rather than the robotic process alone; the platform condition (10.4\%) indicates that the fruit-handling system contributes additional puncture risk.
Taken together, the immediate grading and delayed-inspection results indicate that the harvesting and handling pipeline introduces limited fruit damage, with the dominant effect of the full robotic process being a modest shift from Extra Fancy to Fancy rather than downgrading.

\section{Discussion}\label{sec:discussion}

The field trials demonstrated that the robotic system achieved a harvesting success rate of approximately 80\% and a mean per-arm cycle time of 7.53~s across two commercial orchards with distinct canopy structures.
Despite these encouraging results, several limitations observed during deployment suggest directions for improvement.

The foundation-model-based localization pipeline performs reliably when apples are partially visible, but fails or degrades when a fruit is almost entirely obscured by dense foliage or structural elements such as trellis wires.
In such cases, two primary failure modes arise. First, the point cloud associated with the visible apple surface is too sparse for DBSCAN to form a valid cluster, resulting in a missed detection. Second, when only a small portion of the fruit is visible, the estimated picking point may deviate significantly from the fruit center, reducing the probability of successful attachment even if the arm reaches the target position.
Addressing these cases requires perception models capable of reasoning about partially visible objects---for instance, by incorporating shape priors or fruit-size estimation~\citep{zhu2025foundation} to infer the full fruit geometry from partial observations.
Complementarily, mounting a camera on or near the end-effector would enable visual servoing during the final approach, allowing real-time correction of position estimates.
Beyond improving localization accuracy, the perception system could also estimate the likelihood of successful attachment for each candidate apple and automatically skip targets that are heavily occluded or otherwise mechanically unfavorable, thereby reducing wasted attempts and improving overall throughput.

During operation, the robotic arms inevitably interact with surrounding foliage and branches.
The current Control Barrier Function formulation enforces joint limits that encode self-collision constraints between the arm body and the prismatic stage (Section~\ref{sec:cbf}), but it does not explicitly account for environmental interactions.
This leads to two failure modes: branches displaced during approach may cause the target apple to shift before contact, and branches contacted during retraction can dislodge an already-attached fruit.
A more advanced motion planner would distinguish between compliant canopy elements, such as leaves and flexible side branches that can be safely pushed aside, and rigid structures, such as the main trunk, trellis wires, and wooden stakes, that must be strictly avoided.
The CBF framework naturally extends to such constraints: rigid environmental obstacles can be incorporated as additional barrier functions that keep the end-effector outside exclusion zones, while compliant elements could be handled through soft barrier penalties that bound the allowable contact force rather than prohibiting contact entirely.

A subset of harvest failures is attributable to the vacuum system rather than perception or trajectory errors.
Leaves interposing between the end-effector and the apple surface prevent seal formation, while apples with strong stems or unfavorable orientations resist detachment even when suction is established; the rotational detachment action introduced in Section~\ref{sec:single_arm_cycle} mitigates the latter but cannot eliminate it entirely.
The shared single-vacuum architecture is a deliberate design trade-off that reduces cost and weight; the coordination strategy presented in Section~\ref{sec:coordination} maximizes its utilization, but the hardware constraint remains.
One promising direction is selective harvesting: rather than targeting all reachable apples indiscriminately, the system could prioritize fruit that are ripe and mechanically ready for detachment~\citep{zhu2025foundation}, skipping apples likely to resist the vacuum and returning to them later or leaving them for a subsequent pass.

In the current deployment, a human operator drives the tractor and performs platform adjustments at each harvesting stop.
The vertical arm configuration simplified this process relative to our previous horizontal system~\citep{zhu2025advancement}---each stop now targets a single tree and requires only a small forward-backward translation and a minor vertical adjustment---but the operation remains manual and inherently limits throughput.
Replacing the tractor-trailer with an autonomous ground vehicle would enable continuous low-speed traversal of orchard rows, with the platform repositioning dynamically according to fruit distribution rather than halting at fixed intervals.
Reduced arm weight would further support this transition, lowering the vehicle payload requirement and enabling faster arm motion with less canopy disturbance.

The current system employs a single dual-arm harvesting module.
In this deployment, the control pipeline---including trajectory generation, CBF safety filtering, and dual-arm coordination---runs on an NVIDIA Jetson AGX Orin 64GB Developer Kit, while the perception pipeline remains on a separate workstation to leverage its higher computational capacity and ensure robust real-time inference of the foundation models. This split architecture proved reliable during the field trials, but tethers each harvesting module to a dedicated high-performance computer.
We envision a configuration in which several dual-arm modules are mounted symmetrically on both sides of an autonomous vehicle, allowing a single pass along an orchard row to harvest both sides simultaneously.
Within each side, stacking multiple modules vertically would provide continuous coverage of the canopy height, eliminating the need for repeated vertical platform repositioning.
Consolidating the full software stack---including perception---onto edge-computing devices such as the Jetson is a prerequisite for this multi-module architecture, as each module must operate as a self-contained unit without relying on a central workstation.
Combined with autonomous navigation and the coordination strategy presented in this work, such an architecture could operate as a fully autonomous, scalable harvesting system capable of meeting the throughput demands for commercial apple harvesting.

\section{Conclusion}\label{sec:conclusion}
This paper presented a modular dual-arm apple harvesting robot with a vertical arm configuration designed for commercial orchard deployment. The system integrates foundation-model-based perception, jerk-bounded trajectory generation with CBF safety filtering, a linear sweep harvest strategy with rotational detachment, and a temporal-logic-based dual-arm coordination policy with vision--arm asynchronous scheduling.

Field trials conducted at two commercial orchards in Washington State during the 2025 harvest season demonstrated an 80.0\% per-attempt success rate across 1,738 arm cycles, with a mean per-arm cycle time of 7.53\,s at 95\% actuation capacity. Compared to our previous horizontal dual-arm system~\citep{zhu2025advancement}, the vertical configuration simplified platform positioning from multi-tree lateral repositioning to single-tree stops, and the improved trajectory controller reduced per-arm cycle time by 9.2\%. Indoor coordination experiments confirmed that the proposed scheduling strategy achieves a system throughput of 3.65\,s per attempt---a 1.76$\times$ speedup over serialized operation and within 3.7\% of the unconstrained dual-pump upper bound. To contextualize these results, \citet{yan2025intelligent} reported up to 72.4\% parallel utilization with workspace zoning for dual-arm harvesting, while \citet{li2023multi} demonstrated a four-arm platform with reinforcement-learning-based coordination; the approach presented here achieves near-optimal utilization of a single shared vacuum source through formal temporal-logic specification rather than learned policies. Fruit damage assessments showed that 91.2\% of robotically harvested fruit retained the highest USDA grade (Extra Fancy), with bruise rates between 2.4\% and 4.9\%, confirming that the harvesting and handling pipeline preserves fruit quality under field conditions.

The primary limitations observed during deployment include perception degradation under heavy foliage occlusion, the absence of environment-aware motion planning for canopy interaction, and the reliance on a human operator for platform positioning. Future work will focus on integrating visual servoing for close-range localization refinement, environment-aware trajectory planning that distinguishes compliant canopy elements from rigid structures, and autonomous navigation to enable continuous harvesting without manual repositioning.

\vspace*{\fill}
	
\subsubsection*{Acknowledgements}
This project was funded by the USDA-SCRI project ``AIMS for Apple Harvest and In-Field Sorting'' (Project No. 2023-51181-41244). The findings and conclusions in this paper are those of the authors and should not be construed to represent any official USDA or U.S. Government determination or policy. Mention of commercial products in the paper does not imply endorsement by USDA over those not mentioned.

\bibliography{reference}
	
\newpage
	
	
\newpage
	
	
\end{document}